\documentclass[11pt]{article}

\usepackage[preprint]{acl}

\usepackage{times}
\usepackage{latexsym}

\usepackage[T1]{fontenc}

\usepackage[utf8]{inputenc}

\usepackage{microtype}

\usepackage{inconsolata}

\usepackage{graphicx}

\newtheorem{theorem}{Theorem}[section]

\newtheorem{definition}[theorem]{Definition}

\newtheorem{example}{Example}[section]

\newcommand{\g}{\ensuremath{\mathcal{G}}}
\newcommand{\V}{\ensuremath{\mathbf{V}}}
\newcommand{\E}{\ensuremath{\mathbf{E}}}
\newcommand{\Z}{\ensuremath{\mathbf{Z}}}

\usepackage{tikz}
\usepackage{subfigure}
\usepackage{amsmath}
\usepackage{algorithm}
\usepackage{algorithmic}
\usepackage{amssymb}
\usepackage{listings}

\title{Consistency evaluation of benchmarks used for causal discovery}

\author{Yuzhe Zhang\thanks{Contacts: Yuzhe Zhang: zhangyuzhe1212@gmail.com,\\ Chihui Chen: aryachihuichen@gmail.com,\\ Lina Yao: lina.yao@unsw.edu.au,\\ Chen Wang: chen.wang@csiro.au.} \\
  Independent researcher \\
  Australia \\\And
  Chihui Chen \\
  UNSW \\
  Australia \\\And
  Lina Yao \\
  UNSW \\
  Australia \\\And
  Chen Wang \\
  CSIRO \\
  Australia \\}

\begin{document}
\maketitle
\begin{abstract}
In graphical causal model, causal discovery aims to construct a causal graph based on numerical data and domain knowledge in plain text.
However, the evaluation of causal discovery methods remains a challenge in the area as the progress of domain researches often makes benchmark causal graphs contain mis-aligned knowledge.
This problem especially affects the evaluation of large language model (LLM) based causal discovery methods as they are sensitive to the new discoveries in the literature.
This work is the first to systematically study the quality of benchmark causal graphs.
Specifically, we design a pipeline that automatically retrieves relevant research papers from scientific databases, and prompts LLMs to check the consistency between the benchmark causal graphs and domain research papers.
We evaluate 11 popular real-world benchmarks, for which our pipeline in total proceeds 38,081 domain papers.
Our results show that popular benchmarks vary significantly in their consistency with domain research, with clear implications for causal discovery research.
\end{abstract}

\section{Introduction}
\label{sec:intro}
Causal discovery \cite{glymour2019review} aims to identify the causal links between variables, being a fundamental direction in both artificial intelligence \cite{zheng2018dags,roy2025causal}, and application domains such as medical science \cite{petersen2010alzheimer}, biology \cite{sachs2005causal}, and other fields \cite{huang2021benchmarking}.

Our work considers the graphical model-based causal discovery problem for time-independent datasets.
The problem's target is to construct a causal graph, represented as a (mixed) directed graph, based on numerical data (statistical-based methods) or domain expertise.
Numerous methods have been proposed to accomplish such tasks, e.g., PC \cite{spirtes2001}, GAS \cite{chickering2002optimal}, and more recent ones such as NoTears \cite{zheng2018dags}.
These algorithms achieve impressive performances, evaluated against synthetic or real-world benchmarks.
However, a crucial dilemma of evaluation benchmarks hides these methods' effectiveness in handling real world challenges:
A high-quality real-world benchmark dataset should contain a convincing ground truth causal graph, but this is challenging for many domains.
As a result, many emerging causal discovery works rely on old real-world benchmarks, e.g., \cite{lauritzen1988local,spiegelhalter1992learning}, or semi-synthetic benchmarks, e.g., \cite{tu2019neuropathic}, in their evaluation process.
\cite{gentzel2019case} and \cite{brouillard2025landscape} reveal this issue in the causal discovery domain, and advocate building and using new benchmarks.

Recent works, e.g., \cite{zhang2024causal}, reveal that some of the popular real-world benchmarks contain out-of-date domain knowledge, potentially producing evaluation results that are inconsistent with the truth.
This problem especially undermines the recently exploding new streamline of large language model (LLM)-based causal discovery methods.
Unlike statistical-based methods that heavily rely on the alignment between the numerical data and the ground truth causal graph, LLM-based methods explore knowledge in a ``world model''~\cite{hao2023reasoning} to construct the causal graph.
Therefore, using such practically non-aligned benchmarks for LLM-based methods especially drives the evaluation in an undesirable direction.

\paragraph{A motivation example}
The benchmark Asia \cite{lauritzen1988local}, shown in Figure \ref{fig:asia}, is a popular causal discovery benchmark which is used many LLM-based works \cite{shen2025exploring,vashishtha2025causal,jiralerspong2024efficient,takayama2024integrating}.
These LLM-based methods achieve high performances against the Asia benchmark, usually fully/almost recover the causal graph.
However, domain research suggests that many of the benchmark's causal/conditional association relationships need update.
For instance, the deep learning technique can now reveal bronchitis via x-ray scan \cite{chen2020diagnosis,ntiamoah2021recycling,nishino2014practical}, contradicting against the benchmark, e.g., X-ray and bronchitis are mutually independent conditioned on smoking.
Therefore, a high evaluation performance against such an out of date benchmark contrarily reveals problems of the method in practice.
\medskip

In this work, we intend to examine the quality of popular real-world causal discovery benchmarks.
First, we review the time independent causal discovery works published in artificial intelligence venues in recent years, and summarise the real-world benchmark usage in these works.
As the benchmark issue strongly influences LLM-based causal discovery methods, we also show the trend of this emerging stream based on the meta-data from the arXiv.
The number of LLM-based works drastically increases since 2023, expressing the necessity of systematically evaluating the causal discovery benchmarks.

Second, we investigate the alignment between the popular causal discovery benchmarks and corresponding domain research.
The local Markov property of causal graphs entails a series of conditional association relationships between variables (the d-separation concept), which provide a natural connection to the controlled experimental analysis in most of the domain research works.
We design a LLM-equipped pipeline to verify the consistency between the causal graphs and domain research papers.
Practically, we evaluate 11 real-world benchmarks which are the most frequently used among the surveyed 69 benchmarks, and we analyse 38,081 research papers to mine the scientific evidence that supports or contradicts against the benchmarks.

Our results show that these popular benchmarks vary in their alignment with the corresponding domain research.
Some popular benchmarks have a high quality, e.g., the Sachs benchmark, whilst others present high misalignment with the domain knowledge, e.g., the Child, Asia, and Barley.
These can be a guide for causal discovery works to better select benchmarks in their evaluation process.

\noindent {\bf Related works}
Several papers, e.g., \cite{brouillard2025landscape,gentzel2019case}, review the benchmark usage in causal discovery works, and summarise the benchmark types (i.e., synthetic or practical).
Further before, \cite{glymour2019review} encourages using semi-synthetic benchmarks to better align with the practice.

To the best of our knowledge, we are the first to systematically evaluate common benchmarks from the angle of domain research alignment.
Our benchmark survey updates benchmark usage in the causal discovery domain, especially, benchmarks' popularity and usage in the emerging LLM-based causal discovery works.
We postpone the detailed related work section to Appendix \ref{sec:add_related_work}.


\section{Causal discovery benchmark revisit}\label{sec:review}
We focus on semi-synthetic and real-world causal discovery benchmarks.
Such datasets are constructed based on either human annotations, e.g., \cite{tu2019neuropathic}, statistical based methods using human selected features, e.g., \cite{spiegelhalter1992learning}, or a combination of the two types of methods, e.g., \cite{sachs2005causal}.
As human judgment plays an important role in constructing such datasets, we call them empirical benchmarks.
We evaluate the alignment with domain research of the popular benchmarks in Section \ref{sec:evaluation}.

This section surveys empirical benchmarks used in causal discovery papers published in the recent major AI conferences, and characterize the trend of LLM-based causal discovery papers by mining the meta-data retrieved from the arXiv repository.


\subsection{Benchmark usage in AI publications}\label{sec:benchmark_review}

\paragraph{Paper selection}
We select papers published in main AI conferences in the latest 5 years. These conferences include NeurIPS, ICML, ICLR, AAAI, IJCAI, ACL, and EMNLP.
The selection process consists of two rounds.
In the first round, we filter published papers using keyword ``causal", and manually select the papers that propose methods for {\em time-independent causal discovery}, based on papers' full content.
In the second round, we check the benchmarks used in each paper's experimental, remove papers that purely use synthetic datasets, and record the empirical benchmarks used in the remaining papers.
Note that we do not include papers that study causal relationship in language itself or among multimodal data. 


\paragraph{Summary}
\begin{figure}[t]
    \centering
    \begin{subfigure}
        \centering
        \includegraphics[width=0.5\textwidth]{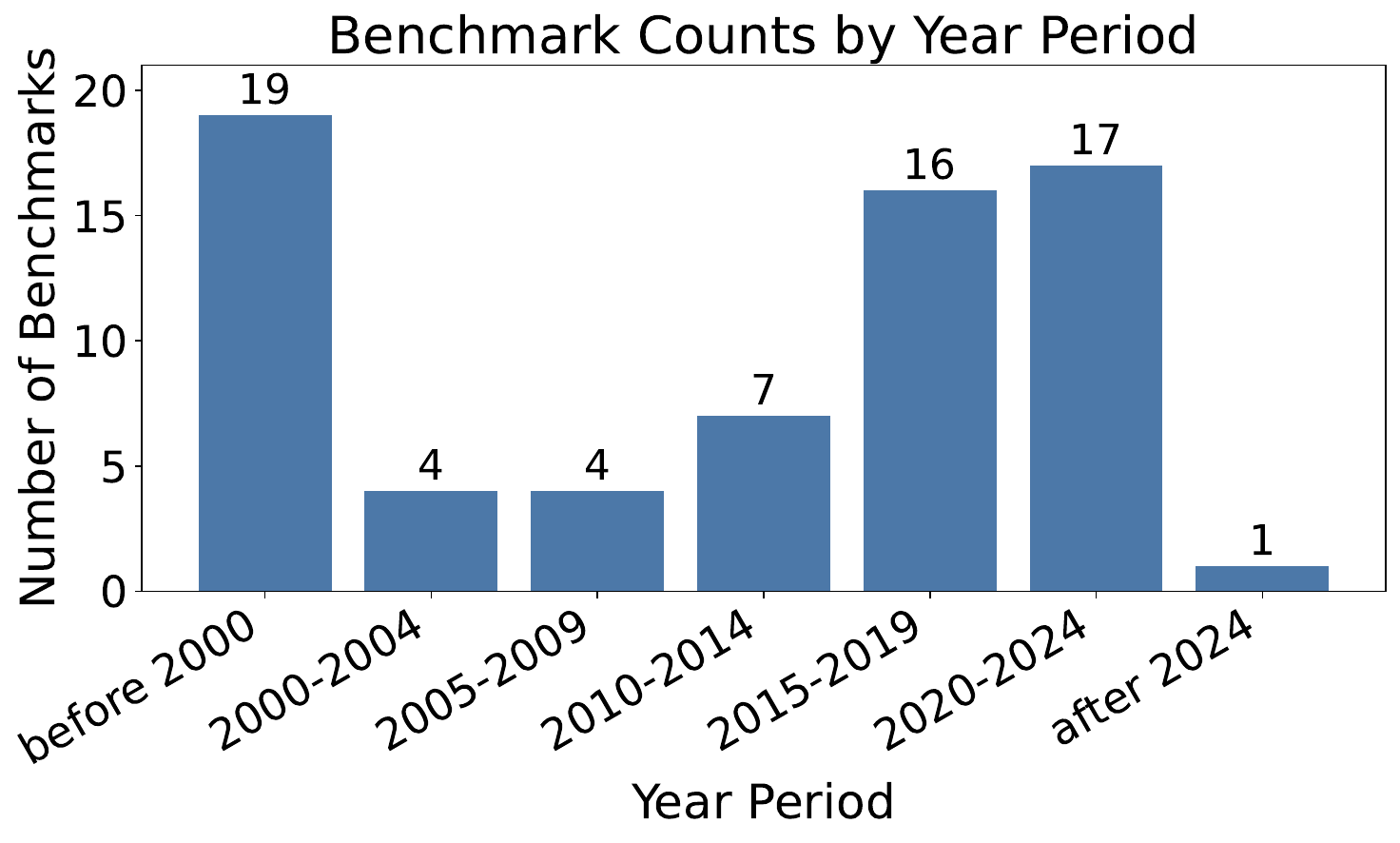}
        \caption{All benchmarks by year period.}
        \label{fig:year-period-all}
    \end{subfigure}
    \hfill
    \begin{subfigure}
        \centering
        \includegraphics[width=0.5\textwidth]{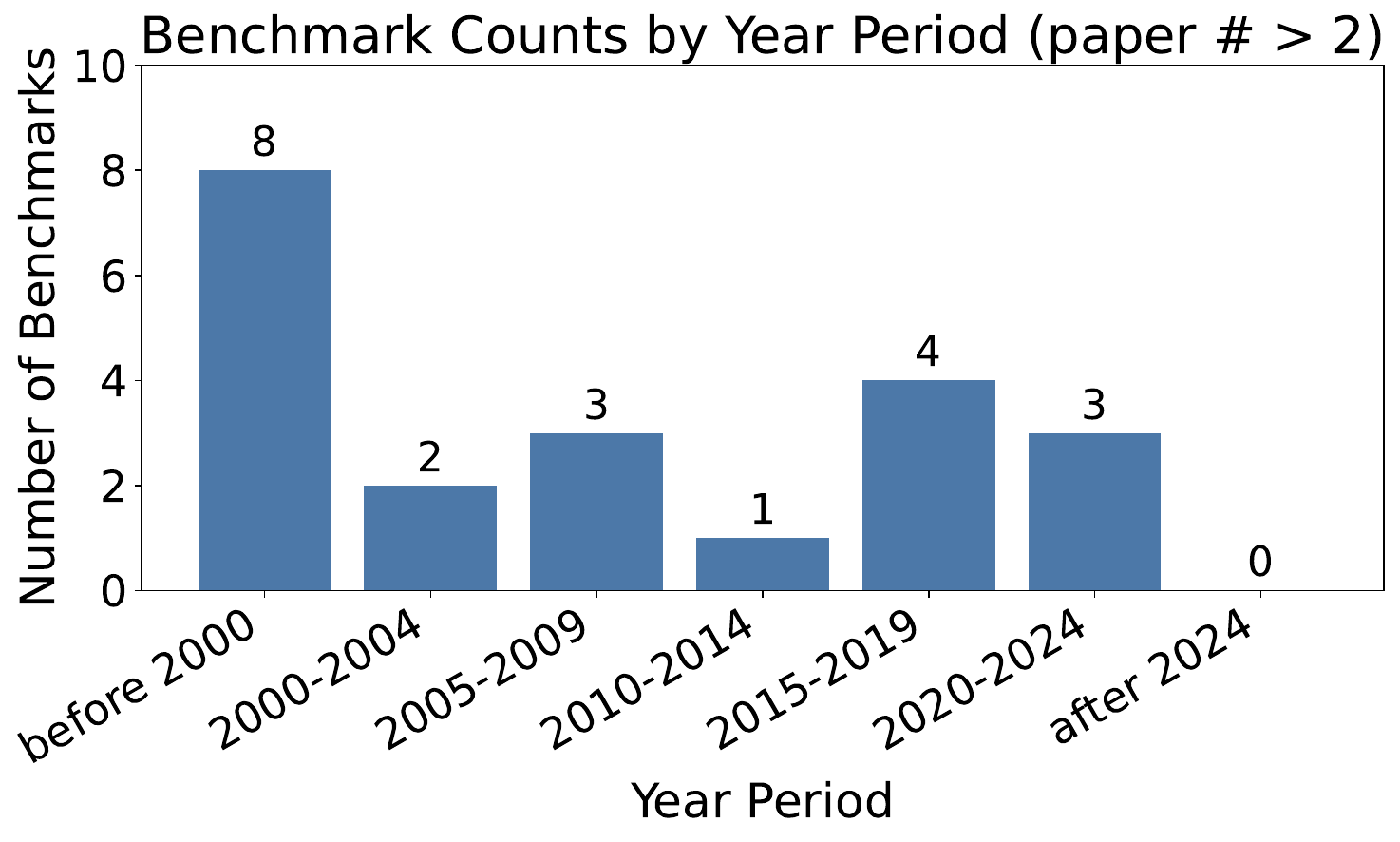}
        \caption{Benchmarks used in at least 3 papers by year period.}
        \label{fig:year-period-filtered}
    \end{subfigure}
    \hfill
    \label{fig:year-period-comparison}
\end{figure}
\begin{figure}[t]
    \centering
    \includegraphics[width=0.5\textwidth]{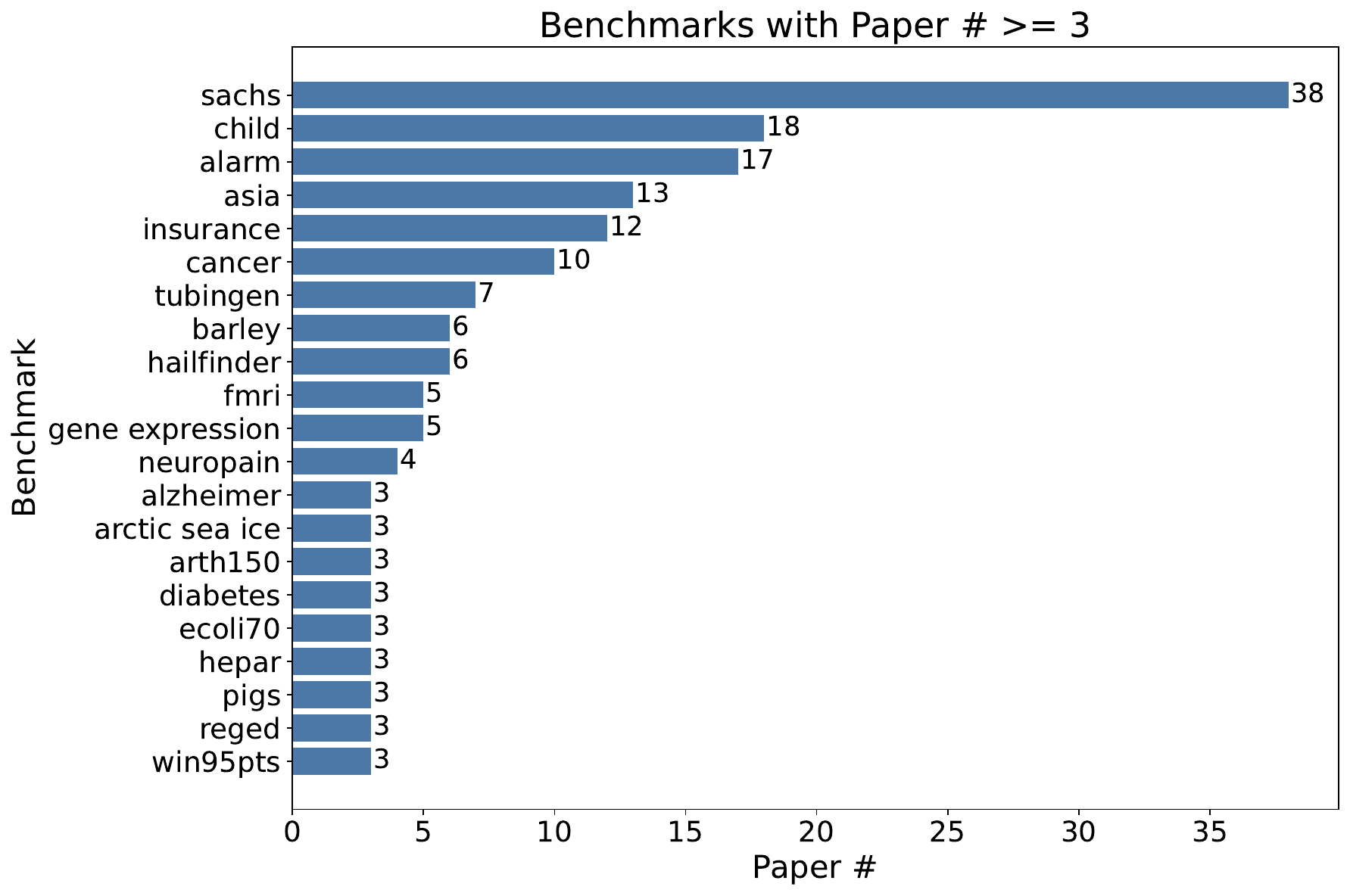}
    \caption{Number of papers that use frequently used benchmarks.}
    \label{fig:frequent_benchmarks}
\end{figure}
The selection process results in 96 papers with at least one empirical benchmark. There are in total  69 unique benchmarks used in these papers.
The benchmark-paper details are listed in Appendix \ref{sec:benchmark-paper}.

Figure \ref{fig:year-period-all} counts the number of benchmarks cited by these papers of a given year, and Figure \ref{fig:year-period-filtered} counts the number of frequently used benchmarks, i.e., those appear in at least three papers.
Among the 69 benchmarks, 19 were published before 2000, showing that the domain still uses many old benchmarks.
Since 2000, new benchmarks have been increasingly constructed. 
There are 34 new benchmarks emerged since 2015, indicating the community is aware of the insufficient benchmark problem in recent years.

21 out of 69 benchmarks are frequently used benchmarks (Figure \ref{fig:frequent_benchmarks}).
The distribution of the frequently used benchmarks is relatively even over years.
Researchers still like to use many of the old benchmarks, observing 8 out of 21 are constructed before 2000, while 7 out of 21 are after 2014.
The usage frequencies of these 21 benchmarks vary significantly.
Sachs is the most popular benchmark, being used in 38 papers.
All of the top 9 benchmarks, expect the pairwise T\"ubingen dataset, are recorded in the bnlearn network repository\footnote{https://www.bnlearn.com/}, revealing its popularity in the domain.
15 among the 21 popular benchmarks are from the domains of medical science and biology.

\paragraph{LLM-based methods}
LLMs have become a tool in causal discovery in recent years. Among the 96 papers, 10 papers propose LLM-based causal discovery methods.They were published very recently with 9 of them published since 2025.
That is, LLM based methods make 31\% of the surveyed causal discovery papers after 2024 (29 papers in total).
There are 28 distinct benchmarks used in these 10 papers: these papers use diverse benchmarks (see details in Appendix \ref{sec:LLM_papers}).
However, the top 3 frequently used benchmarks in LLM-based papers are: Asia, Cencer, and Sachs, all of them from the bnlearn small network repository.

\subsection{Emerging LLM-based causal discovery works}

\paragraph{arXiv search process}
As new papers often appear in arXiv first, analyzing their benchmark use can reveal the trend in this area. We perform a targeted bibliometric analysis of arXiv papers at the intersection of causal discovery and LLMs.
Specifically, we form a search query by two keyword sets: one for causal discovery and one for LLMs.
Then, we submit the queries to the arXiv API to search papers of which the title or the abstract contain both keywords, expecting to retrieve LLM-based causal discovery papers.

We also extract benchmark usage information contained in the abstract through rule-based matching against an extended benchmark inventory constructed by LLMs (details shown in Appendix \ref{sec:arxiv_benchmark}).



\begin{figure}[t]
\centering
\begin{subfigure}
\centering
\includegraphics[width=\linewidth]{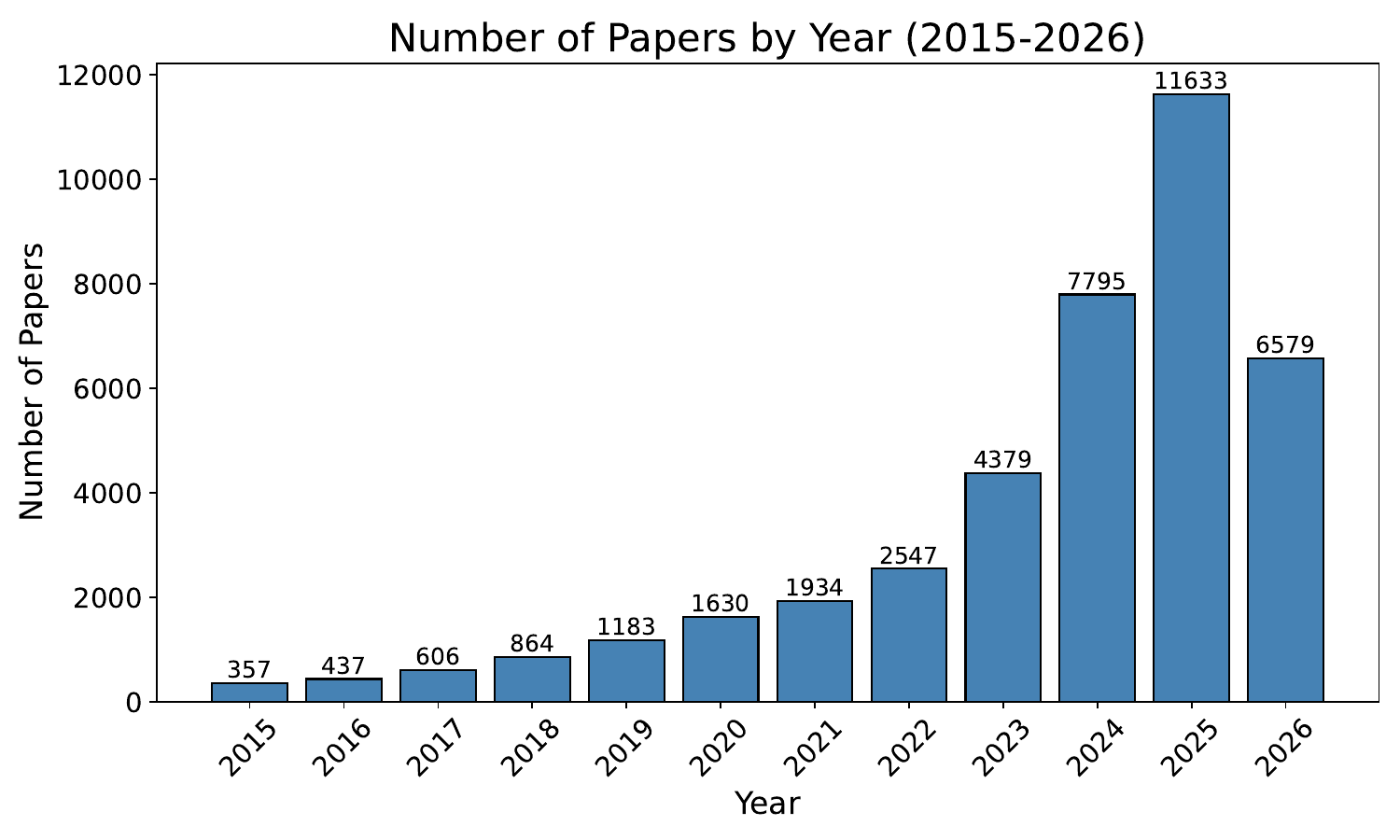}
\caption{Number of papers/year by searching ``causal discovery" and ``LLM".}
\label{fig:arxiv_cd}
\end{subfigure}
\begin{subfigure}
\centering
\includegraphics[width=\linewidth]{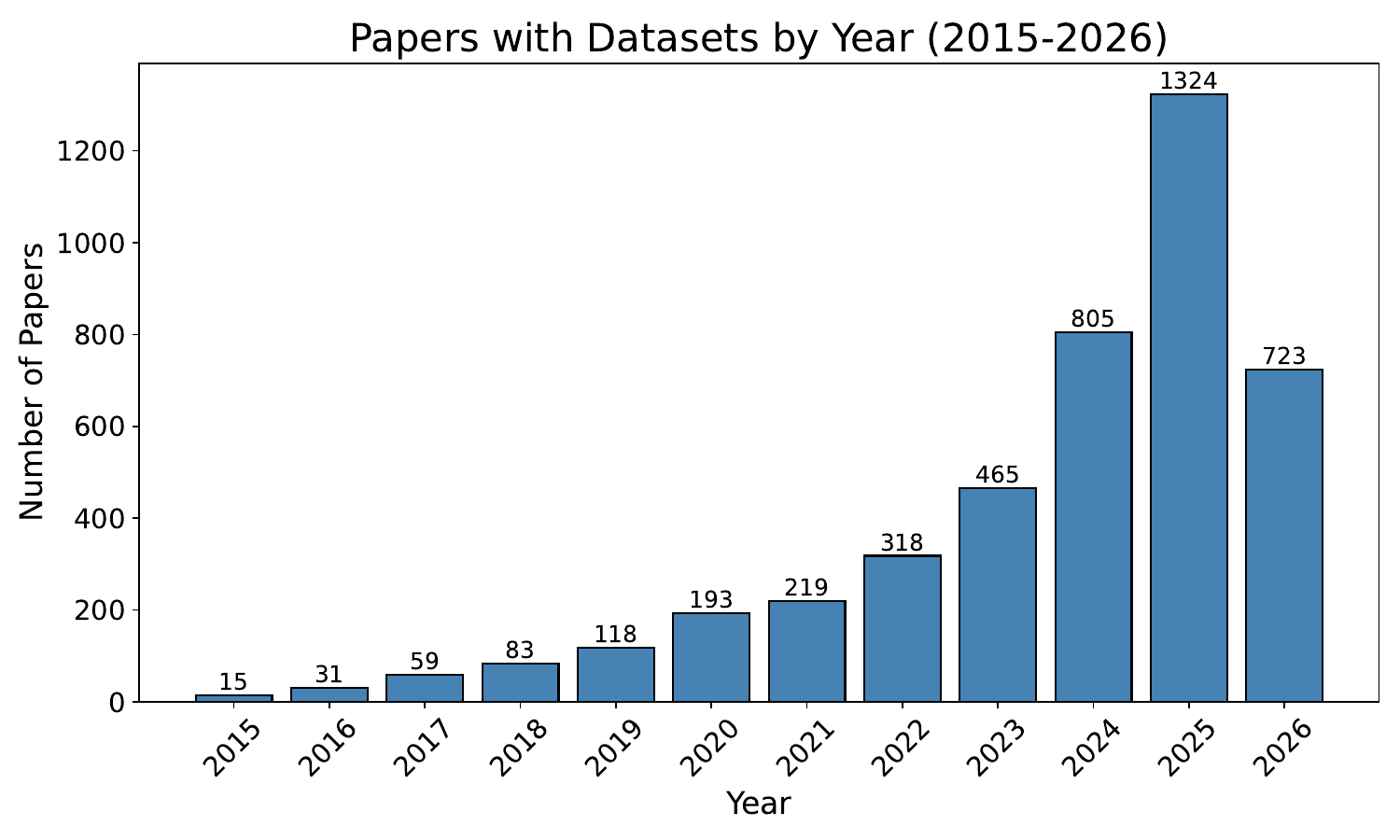}
\caption{Number of papers/year by searching ``causal discovery", ``benchmark", and ``LLM".}
\label{fig:arxiv_cd_bench}
\end{subfigure}
\label{fig:arxiv}
\end{figure}

\paragraph{Summary}
Figures \ref{fig:arxiv_cd} and \ref{fig:arxiv_cd_bench} show the number of papers that propose LLM-based causal discovery methods and LLM-based causal discovery with evaluation by empirical datasets, respectively.
The two numbers have a similar trend: both of them stably increase from 2015 to 2025, and the trends have become drastic since 2022.
This trend is strongly stimulated by the rapid advancement of LLMs in recent years.
In 2025, there are up to 11633 LLM-based causal discovery papers, among which up to 1324 papers contain evaluation based on empirical benchmarks.

In Appendix \ref{sec:arxiv_benchmark}, the arXiv search results show the top 20 benchmarks and our survey benchmarks have overlapped benchmarks: Link, Cancer, Water, and Child.
Three of them are recorded in the bnlearn repository, showing its popularity.

\section{Domain knowledge consistency evaluation}\label{sec:evaluation}

\subsection{Preliminary: graphical causal models}
\paragraph{Causal graph}
A causal graph is a directed acyclic graph (DAG), denoted as $\g=\langle \V, \E \rangle$, where $\V$ is the set of variables, $\E\subseteq \V\times \V$ is the set of directed edges.
For each pair of variables $V_i, V_j\in \V$, $(V_i, V_j)$ presenting in $\E$ denotes an directed edge from $V_i$ to $V_j$ exists, showing that $V_i$ is a {\em direct cause} of $V_j$.
An ordered sequence of distinct variables $(V_{i1}, V_{i2},\cdots, V_{ik})$ from $\V$ forms a path if each adjacent pair $V_{ij}, V_{i(j+1)}$ are adjacent in $\g$ where $1\le j\le k-1$, and the path is a directed path if $(V_{ij}, V_{i(j+1)})$ is in $\E$.
Typically, a DAG does not contain any cycle, i.e., a path with the first and the last variables coinciding.
For each variable $V_i\in \V$, let $pa_\g(V_i)$ denote the parents of $V_i$ in $\g$, i.e., $pa_\g(V_i) = \{V\in \V\mid (V, V_i)\in E\}$, and let $de_\g(V_i)$ (resp. $nd_\g(V_i)$) denote the descendants (resp. non-descendants) of $V_i$ in $\g$, i.e., all variables that are reachable (resp. unreachable) through a directed path from $V_i$.
For a triple $V_i, V_j, V_k\in \V$, such that $V_i\rightarrow V_j \leftarrow V_k$ and $V_i$ and $V_k$ are not adjacent in $\g$, we call $V_j$ a collider.

\paragraph{Local Markov property}
An ubiquitous assumption for a causal graph is the local Markov property.
That is, for each variable $V\in\V$, it holds that:
$$V\perp nd_\g(V) \mid pa_\g(V),$$
where $\perp$ denotes statistically independent.

\paragraph{d-separation}
The local Markov property entails a series of conditional association relationship between variables.
The definition of {\em d-separation} characterises such conditional association relationship reflected from a causal graph.

\begin{definition}[d-separation]
For a pair of variables $V_i, V_j\in \V$, a variable set $\Z\subseteq \V\setminus \{V_i, V_j\}$ d-separates $V_i$ and $V_j$ if the following hold:\\
(1) $\Z$ contains at least one variable on each collider free path between $V_i$ and $V_j$; and\\
(2) If a path between $V_i$ and $V_j$ contains a collider, the collider and its descendants are not contained in $\Z$, or the path is blocked by a non-collider node on the path.\\
If $\Z$ d-separates $V_i$ and $V_j$, then, it holds that:
$$V_i \perp V_j \mid \Z.$$
\end{definition}

In a DAG causal graph, d-separation specifies all conditional association relationships of any triple $V_i, V_j$ and $\Z\subseteq \V\setminus\{V_i, V_j\}$: if $\Z$ d-separates $V_i$ and $V_j$, the pair of variables are mutually independent conditioned on $\Z$, otherwise they are correlated.
Note that $\Z$ can be empty.

\begin{example}\label{ex:asia}
Figure \ref{fig:asia} shows the causal graph of the Asia benchmark dataset from the BNLearn repository \cite{scutari2019package}.
Observe that variables $Xray$ and $Bronchitis$ are d-separated by $Smoking$, and therefore, we have that
$$Xray \perp Bronchitis \mid Smoking.$$
On the other hand, if two variables are adjacent, e.g., $Smoking$ and $LungCancer$, they cannot be d-separated by any variable set, and therefore, they are always correlated, even conditioned on any other variable set.
\end{example}
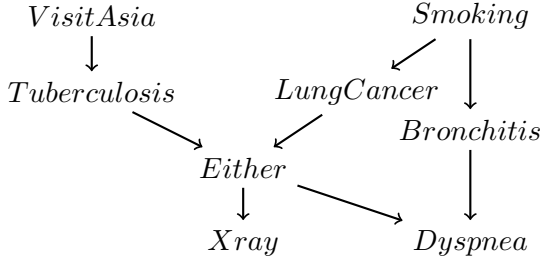
\begin{figure}[t]
\centering
\begin{tikzpicture}
\node (as) at (-4,2) {$VisitAsia$};
\node (tb) at (-4,1) {$Tuberculosis$};
\node (ei) at (-2,0) {$Either$};
\node (sm) at (1,2) {$Smoking$};
\node (lc) at (-0.5,1) {$LungCancer$};
\node (xr) at (-2, -1) {$Xray$};
\node (br) at (1,0.5) {$Bronchitis$};
\node (dy) at (1,-1) {$Dyspnea$};

\path[->,thick] (as) edge (tb);
\path[->,thick] (tb) edge (ei);
\path[->,thick] (lc) edge (ei);
\path[->,thick] (ei) edge (xr);
\path[->,thick] (ei) edge (dy);
\path[->,thick] (sm) edge (lc);
\path[->,thick] (sm) edge (br);
\path[->,thick] (br) edge (dy);
\end{tikzpicture}
\caption{The causal graph of the Asia benchmark dataset: a semi-synthetic benchmark dataset.}
\label{fig:asia}
\end{figure}

\paragraph{Our benchmark evaluation target}
The main objective of this work is to evaluate the consistency of popular benchmarks used in causal discovery publications with the corresponding domain research.
A typical original domain research yields a conclusion by designing a controlled experiment, i.e., defining the target variable pair (e.g., studying whether variable $V_i$ has any influence on corresponding observation of variable $V_j$).
A convincing controlled experiment usually specifies a set of covariates to avoid bias.
With data collection and statistical/machine learning analysis, the conclusion reflects the association relationship between the target variable pair conditioned on the covariates.
This naturally connects with the d-separation concept, providing an interface to verify the consistency between a domain research work and a causal graph.

Specifically, for a causal graph, we systematically retrieve research papers from scientific databases that are relevant to the conditional association relationships (determined by the d-separation criteria) reflected from the causal graph, and then, evaluate what the proportion of the retrieved papers contain inconsistent conclusions than the causal graph's conditional association relationships.
This inconsistency rate measures the mis-alignment between the causal graph and the domain research.

\subsection{Methodology}
We design a LLM equipped pipeline to evaluate causal graphs' consistency with domain research.
The pipeline is composed of three components: (1) domain research paper retrieval, (2) conditional association relationship extraction, and (3) consistency verification.
We mainly utilise LLMs to verify the consistency between paper conclusions and causal graph conditional association relationship in component (3).
Given a causal graph $\g$, for {\em each pair} of variables $V_i, V_j\in \V$, 
the pipeline executes as follows.

\paragraph{Domain research paper retrieval}
The first component constructs a research paper pool where all papers study the relationship between $V_i$ and $V_j$, by searching and downloading papers from a scientific database, and extracting text from each paper.
Basically, we send a query ``$V_i$ and $V_j$'' to search relevant papers in a database, download the accessible papers among the search result, and extract the text from the downloaded files to make papers LLM-reading ready.

A simple search query ``$V_i$ and $V_j$'' is not sufficient for efficient search due to the following facts:
(i) Inconsistent term using is normal in various research domains, e.g., the term ``Smoking'' in Example \ref{ex:asia} may be expressed as ``Tobacco use'' or ``Cigarette consumption''.
(ii) Many database uses keyword matching against the full paper in the search engine, and the search query may retrieve a large number of irrelevant papers.
Hence, we employ the {\em query expansion} \cite{carpineto2012survey} technique to form a search query, and add an extra filter process to screen out irrelevant papers.

The query expansion finds a set of synonyms for each of $V_i$ and $V_j$, i.e., $\{V_i^1, \cdots, V_i^{ki}\}$ and $\{V_j^1, \cdots, V_j^{kj}\}$, and then, it forms a query:
\begin{align*}
\begin{split}
&(V_i^1\text{ OR }V_i^2\text{ OR }\cdots\text{ OR }V_i^{ki})\text{ AND }\\&(V_j^1\text{ OR }V_j^2\text{ OR }\cdots\text{ OR }V_j^{kj}).
\end{split}
\end{align*}
Upon the returned search results, the filter screens in papers of which the abstract contains terms from both sets $\{V_i^1, \cdots, V_i^{ki}\}$ and $\{V_j^1, \cdots, V_j^{kj}\}$.
Then, in the final paper list, we download the pdf files of all open-access papers by interacting with the database's API, and convert the downloaded files to json files.


\paragraph{Conditional association extraction}
Algorithm \ref{algo:association} deterministically extracts the conditional association relationships from causal graph $\g$: (i) if $V_i$ and $V_j$ can be d-separated by an empty set, $V_i \perp V_j$; (ii) if $V_i$ and $V_j$ can be d-separated by set $\Z$, $V_i\perp V_j | \Z$; and (iii) if $V_i$ and $V_j$ are adjacent in $\g$, they are always correlated.
Observe that $V_i$ and $V_j$ can be d-separated by multiple sets, and we record all such conditional association relationships.
Then, for each conditional association relationship, we formulate a query, e.g., ``Is $V_i$ and $V_j$ are mutually independent conditioned on variables in $\Z$?''


\paragraph{Consistency verification}
With the paper contents and the conditional association queries prepared, we employ LLMs to verify if the each paper's conclusion is consistent with each conditional association query.
Especially, we provide the mathematical intuition and example expressions of the each conditional association type in the prompt, and query the LLMs to verify the consistency (see prompt in Appendix \ref{sec:app:process}).
Each LLM query may return ``unknown'' when the paper content cannot support the decision making, or ``consistent''/``inconsistent'' when the paper content is consistent/inconsistent with the query.

\subsection{Practical implementation}
Among the 69 benchmarks (recall the benchmark survey in Section \ref{sec:benchmark_review}), we only evaluate the popular benchmarks, i.e., those are used in at least 3 causal discovery papers.
There are 21 popular benchmarks in total, and we exclude 10 benchmarks: (i) the pairwise benchmark (T\"ubingen) due to its simple conditional association type; (ii)) the large benchmarks (especially, Hailfinder, Gene expression, Neuropain, Arth150, Hepar2, Pigs, Reged, Barley, and WIN95PTS) due to our limited computation capacity.
Table \ref{tab:datasets} lists the details of the evaluated benchmarks.

An $n$-variable benchmark has ${n\choose 2}$ combinations of variable pairs.
For each variable pair, we limit the database search engine to only return up to 500 the most relevant papers, and download the open access papers among the searched papers.
In the 11 evaluated benchmarks, some of them have more than 200 combinations of variable pairs, and we randomly sample 200 variable pairs to evaluate.
Therefore, the pipeline processes up to 100000 full papers for each benchmark\footnote{The actual paper number is much fewer than 100000, since (1) some variable pairs have few search results due to their weak relationship, and (2) only a limited number of papers have open access full content.}.

Most of the benchmark causal graphs are from medical science and biology.
For such causal graphs, we use PubMed\footnote{https://pubmed.ncbi.nlm.nih.gov/} as the scientific database to search papers, and we download papers that are available in the open access database PMC.
For causal graphs from the other domains, we use Openalex\footnote{https://openalex.org/} to search papers, and download papers accessible through the institution's subscription databases.
The downloaded pdf files are converted into json files by the Mistral ocr API.

For variables from the medical science and biology, we expand the search query by finding 15 synonyms from both of the two resources: (1) the UMLS system\footnote{https://www.nlm.nih.gov/research/umls/index.html} and (2) ChatGPT.
For variables from other domains, we only use ChatGPT to find 15 synonyms.

We deploy the Qwen3-30B-A3B-Thinking-2507 model on a single H100 GPU with 80GB memory to verify the consistency between each paper content and the d-separation query\footnote{To show the accuracy of the LLM-based consistency verifier, we randomly sample 100 retrieved papers, and manually check the correctness of the answer. The verifier achieves an accuracy of 0.9. See details in Appendix \ref{sec:app:accuracy}}.

\begin{table*}[t]
\begin{tabular}{|l|l|l|l|l|l|}
\hline
 benchmarks & \# nodes & \# edges & \# node pairs & \# retrieved papers & benchmark publication year \\ \hline
sachs   & 11    & 17  & 55  & 4197   & 2005      \\ \hline
child  & 20   & 25   & 190  & 4073   & 1992  \\  \hline 
alarm  & 37   & 46  & 200  & 12736  & 1989    \\ \hline 
asia    & 8  & 8   & 28  & 1413   & 1988  \\ \hline
insurance  & 27  & 52  & 200 & 3408  & 1997 \\ \hline
cancer & 5  & 4  & 10  & 319 & 2010  \\ \hline
fmri  & 6  & 7  & 15  & 2367  & 2015    \\ \hline
alzheimer & 11  & 19  & 55 & 5616   & 2010   \\ \hline
arctic sea ice  & 12  & 40  & 66 & 784   & 2021 \\ \hline
diabetes & 4  & 4  & 6   & 427 & 1991   \\ \hline
ecoli70  & 46  & 70  & 200  & 2741  & 2005    \\ \hline
\end{tabular}
\caption{Details of selected benchmark datasets.}
\label{tab:datasets}
\end{table*}

\paragraph{The main metric: inconsistency rate}
Recall that investigating each paper returns one of three answers: ``Unknown'', ``Consistent'', or ``Inconsistent''.
The answer ``Unknown'' basically reflects that the paper does not contain sufficient information to support the decision of the conditional association query.
Then, for each benchmark, let $\#_{\text{Con}}$ denote the number of papers that present results consistent with the corresponding d-separation query, and in contrast, let $\#_{\text{InCon}}$ denote the number of papers that show results inconsistent with the corresponding d-separation query.
The {\em inconsistency rate} of a benchmark is then defined as:
$$R_{\text{InCon}} = \frac{\#_{\text{InCon}}}{\#_{\text{InCon}} + \#_{\text{Con}}}.$$
\section{Evaluation results}\label{sec:result}
In the entire process, our pipeline processes 38,081 full papers (see details in Table \ref{tab:datasets}). We measure the inconsistency rates of these papers on benchmarks.


\subsection{Inconsistency rates and paper distribution}
\begin{figure}[t]
    \centering
    \includegraphics[width=0.5\textwidth]{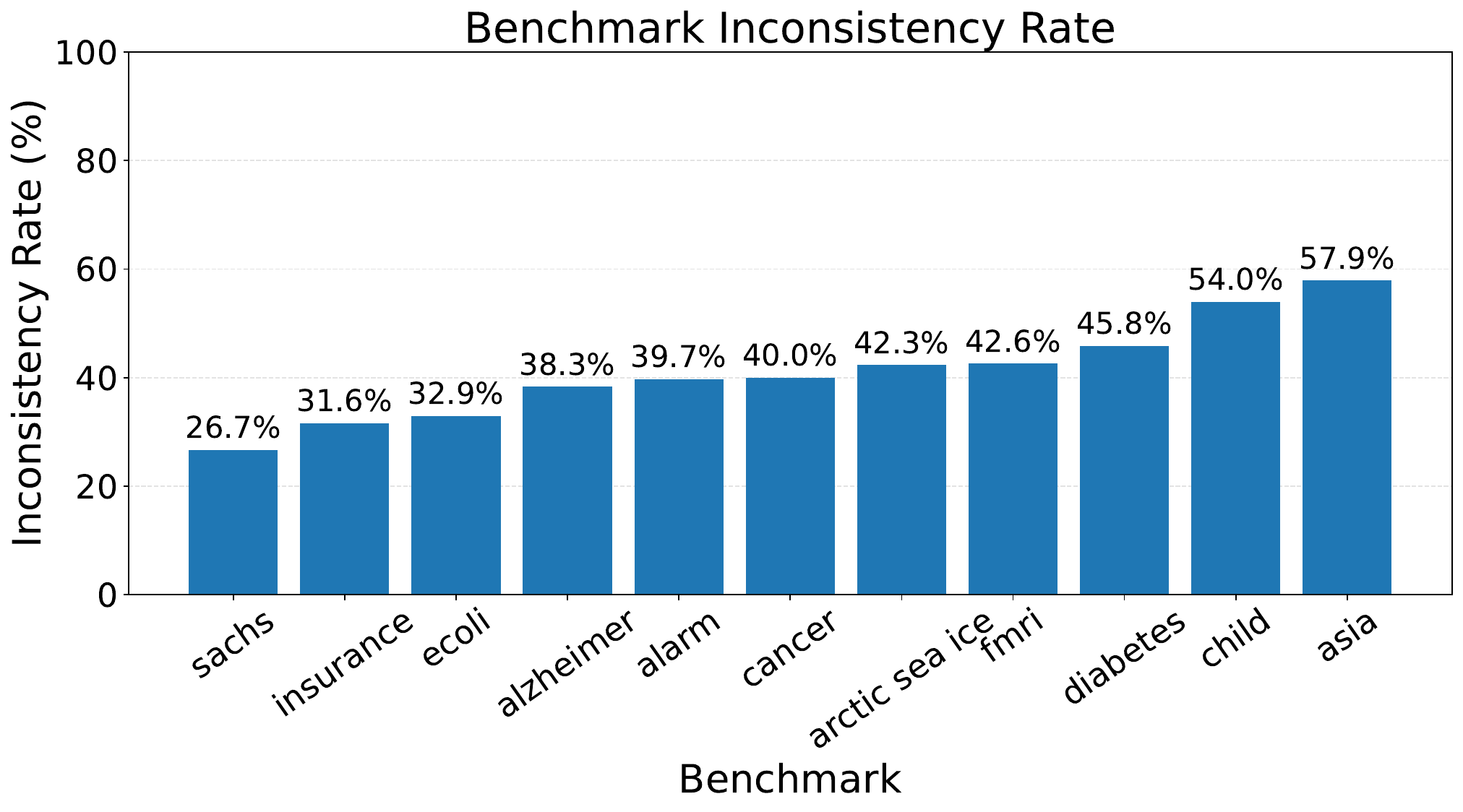}
    \caption{Benchmarks' inconsistency rates.}
    \label{fig:inconsistency_rates}
\end{figure}

\begin{figure}[t]
    \centering
    \includegraphics[width=0.5\textwidth]{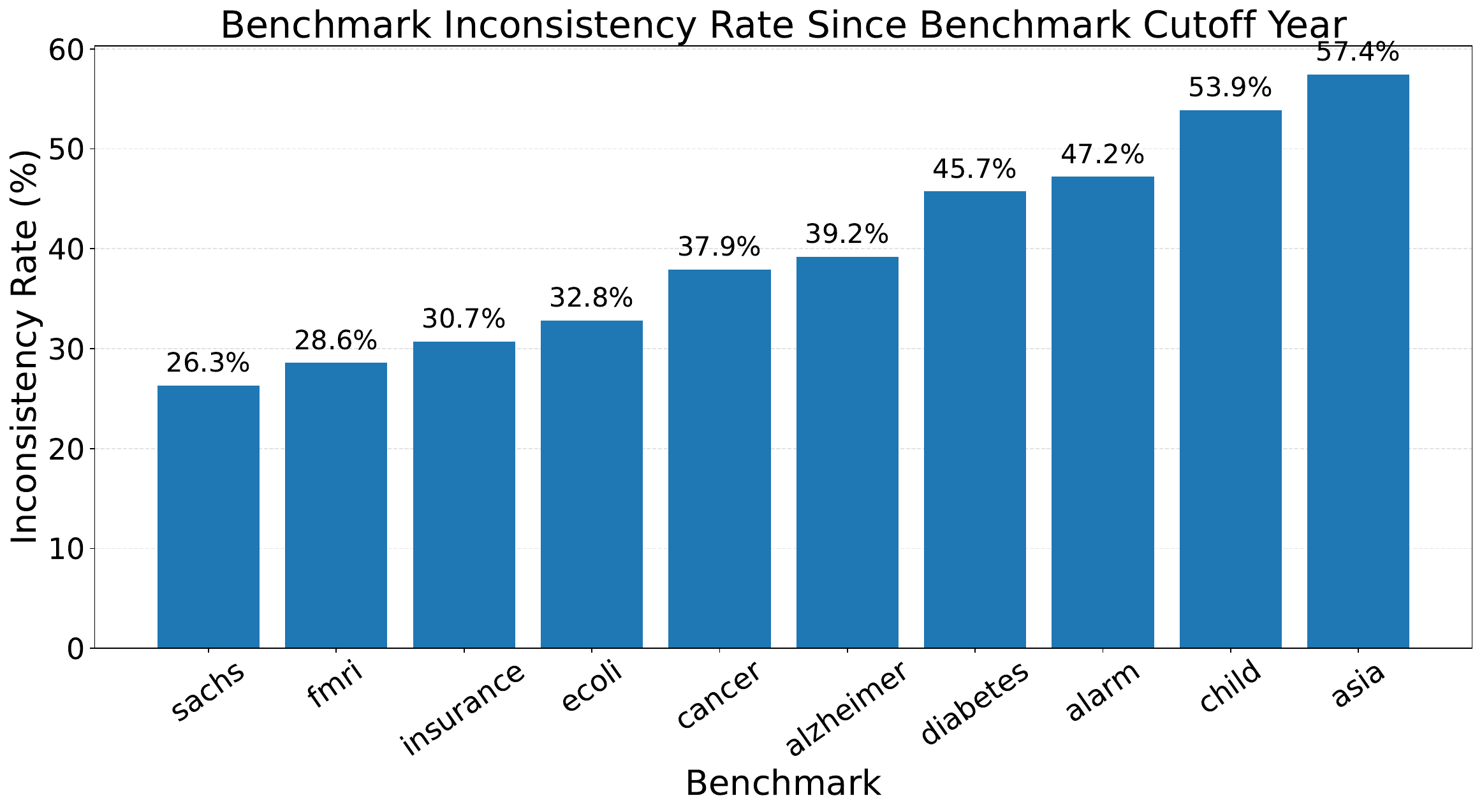}
    \caption{Benchmarks' inconsistency rates since the construction year.}
    \label{fig:inconsistency_rates_since_year}
\end{figure}

Figure \ref{fig:inconsistency_rates} shows the benchmark with the lowest inconsistency rate (26.70\%) is Sachs, which potentially best aligns with the domain research, and it is the most frequently used benchmark (38 surveyed papers use it).
The second and third benchmarks are Insurance and Ecoli, with inconsistency rates 31.6\% and 32.9\%, respectively.

In contrast, Diabetes, Child, and Asia have the highest inconsistency rates (45.8\%, 54.0\%, and 57.9\%, correspondingly).
It is worth noting that all of them are constructed before 2000.

Among 11 benchmarks, only Insurance and Arctic Sea Ice do not belong to the fields of medical science or biology.
However, we do not find the benchmark field is related to benchmarks' inconsistency rates. 

\begin{figure*}[t]
    \centering
    \includegraphics[width=\textwidth]{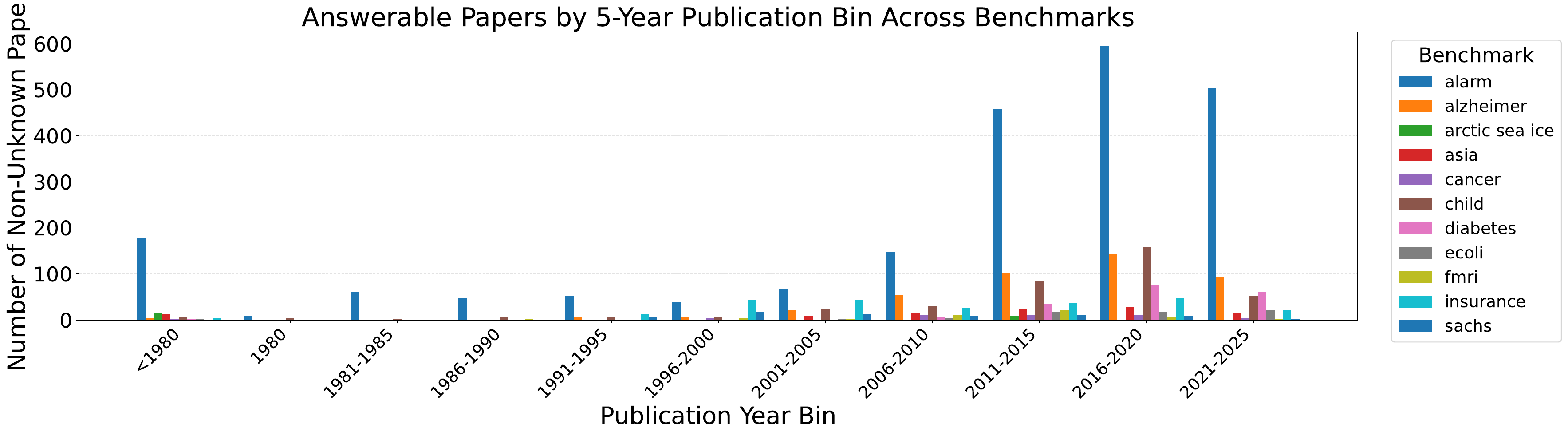}
    \caption{Benchmarks' relevant paper number per year period.}
    \label{fig:answerable_paper_by_year}
\end{figure*}
\begin{figure*}[t]
    \centering
    \includegraphics[width=\textwidth]{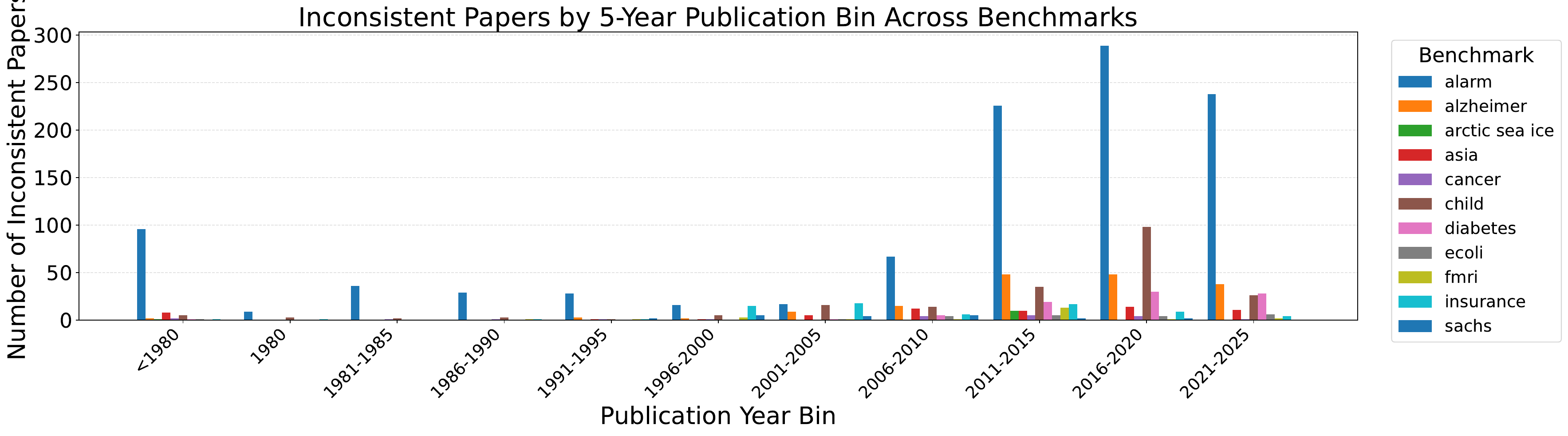}
    \caption{Benchmarks' inconsistent paper number per year period.}
    \label{fig:inconsistent_paper_by_year}
\end{figure*}

We also use the construction year as a cutoff year to measure the inconsistency rate. When only considering papers published after each benchmark's construction year, the inconsistency rates vary significantly as shown in Figure \ref{fig:inconsistency_rates_since_year}.
Fmri's consistency rate reduces from 42.6\% to 28.6\%, with the ranking promotes to the second.
In contrast, Alarm's inconsistency rate increases from 39.7\% to 47.2\%.
All prior year 2000 benchmarks, except for Insurance, are ranked the last, i.e., Diabetes, Alarm, Child, and Asia.

Figures \ref{fig:answerable_paper_by_year} and \ref{fig:inconsistent_paper_by_year} show the number of retrieved relevant papers (i.e., papers do not return ``Unknown'') and inconsistent papers over years, respectively.
Observe that for most of the benchmarks, both paper numbers start increasing since 2001, and achieve the highest from 2011 to 2025.
Interestingly, many of the benchmarks have a relatively lower number of inconsistent papers around their publication year compared to the number in the preceding and subsequent years, especially, Alarm has the most obvious trend (see the detailed inconsistent paper numbers in Appendix \ref{sec:app:paper_count}).

\subsection{Discussion}

\paragraph{Benchmarks' validity over time}
A benchmark often reflects the state-of-the-art understanding of a problem at the time it is constructed. The inconsistency rate may grow along time as the problem understanding progresses. 
Figure \ref{fig:inconsistency_rates_since_year} clearly shows such a trend: dated benchmarks suffer from the domain misalignment problem.
Our results indicate that new benchmarks provide a better measurement for causal discovery methods, especially for LLM-based methods.
On the other hand, maintaining and updating the dated benchmarks help the development of better causal discovery methods.



\paragraph{Hypothesis on growing inconsistency}
For almost benchmarks, the inconsistent paper number increases after benchmarks' publication year, whilst the inconsistency rates do not have such a trend.
We conjecture that the subsequent higher inconsistent paper number is due to richer study on more complex scenarios.
More recently, high-quality datasets with higher dimensions and more advanced data handling techniques have become available.
Therefore, researchers are able to construct more complex causal models, including variables that are latent confounders previously.
Such studies may not invalidate the previous benchmarks, but they consider additional factors missing in previous studies, which may lead to the inconsistency. 

\paragraph{Benchmark construction process matters}
The Sachs benchmark achieves the lowest inconsistency rates in both Figures \ref{fig:inconsistency_rates} and \ref{fig:inconsistency_rates_since_year}.
Though it was constructed over 20 years ago, it retains the highest quality among the evaluated benchmarks.
We conjecture that its construction process mainly contributes to its stable quality.
\cite{sachs2005causal} runs a Bayesian network computational method to reconstruct a basic causal graph.
Then, the work incorporates domain expertise to revise the basic causal graph, and obtain the benchmark causal graph.
Such a hybrid construction process guarantees its alignment with the domain knowledge.
We advocate the domain treat this hybrid process as a standard in benchmark construction.


\section{Conclusion}
We revisited popular benchmarks used in causal discovery works in recent years, and evaluated them based on how they align with changing domain knowledge.
We designed a LLM-powered pipeline to retrieve a large number of research papers, against which the pipeline evaluates the benchmarks.
Our results show that the popular benchmarks have varying alignment levels with the domain research.
Especially, the mis-alignment issue is more severe on old benchmarks.

Future works include improving the paper retrieval performance on both efficiency and accuracy. 
Ultimately, the causal discovery research should develop a systematic method to track knowledge change of existing benchmarks. 

\section*{Limitations}

The limitations of our work is mainly in two folds.

(1) {\bf The accuracy of the LLM paper verifier}
To verify the consistency between a causal graph and each research paper from the vast paper pool, the use of LLMs is a necessity.
However, employing LLMs unavoidably introduces noise in the process.
Empirically, we find that a more advanced LLM tool can significantly improve our evaluation result.
However, this also requires a more generous budget for the computation or LLM API callings.
Due to the vast size of our retrieved paper pool, we deploy a small-size open-source LLM (Qwen3-30B-A3B-Thinking-2507) in the experiment.
We conjecture that the result would be further improved under the use of a more powerful LLM tool.

(2) {\bf Paper retrieving efficiency}
In the current version, we simply send a searching query, which is formulated by a query expansion technique, to scientific databases' search engines.
We aim to obtain a comprehensive and unbiased paper list.
However, the component's paper retrieval efficiency is still low, forcing us to analyse more papers.
Scientific databases also have varying paper retrieving capability.
PubMed performs well on biomedical paper retrieval, however, the other domains rely on different databases, and we find the database quality also bottlenecks our evaluation.

Moreover, the access to full papers is still significantly limited, presenting another main challenge for vast domain paper based experiments such as ours.

\section*{Acknowledgments}
The authors thank Xiang Dai for providing the paper downloading API.


\bibliography{causal}

\newpage
\onecolumn
\appendix

\section{Additional related works}\label{sec:add_related_work}
\paragraph{Causal graph benchmarks}
Evaluation benchmarks are basically divided into three types: purely synthetic, semi-synthetic (synthetic data based on real-world causal graphs), and real-world.
\cite{brouillard2025landscape} provides a summary of benchmark usage of the three types, and reviews the real-world and semi-synthetic benchmarks.
Earlier than this survey, \cite{gentzel2019case} summarises evaluation methods and the usage of different types of benchmarks in the domain.
They advocate using real-world benchmarks in evaluation processes, and choosing interventional data rather than observational data.
Similarly, \cite{glymour2019review} encourages using semi-synthetic data to better align with the practice, compared to using purely synthetic data.

Our work focuses on semi-synthetic and real-world benchmarks.
We summarise the detailed usage of such benchmarks, especially, with an update on the benchmark usage in LLM-based causal discovery methods, and evaluate their aligningness with real-world knowledge.

\paragraph{LLM-based causal discovery}
Compared to statistical-based causal discovery methods, LLM-based methods are particularly vulnerable to benchmarks' non-aligningness with real-world knowledge.
We list several representatives of recent works due to the vast number of new works.

A LLM-based method is either purely LLM-based or hybrid (of LLM-based and statistical-based).
Pure LLM-based methods typically send LLMs pairwise causal relationship queries, e.g., \cite{long2022can}.
\cite{jiralerspong2024efficient} reduces this strategy's complexity by a breadth-first search algorithm.
\cite{zhang2024causal} adds retrieval augmented generation (RAG) to utilise domain research in pure LLM-based method.
\cite{vashishtha2025causal} additionally queries LLMs the causal orders to retain stability.

A hybrid causal discovery method, e.g., \cite{roy2025causal}, combines LLM query and statistical-based method.
\cite{feng2025iris} and \cite{shen2025exploring} use LLMs as a data extractor when tabular data is insufficient.

Moreover, \cite{kiciman2023causal} and \cite{feng2025reliability} examine LLMs' performance in causal discovery, e.g., the reasons of LLMs' success and failure, and LLMs' reliability.

\section{Additional information of literature review}\label{sec:benchmark-paper}


The follows are all benchmark datasets we reviewed in the causal discovery publications.
Each item is a dataset, where the ``original paper'' is the resource paper of the dataset, and the ``papers using it'' element lists all causal discovery publications that use it in the evaluation process.
Note that the original paper might not be the only actual resource paper, for instance, for the Gene expression dataset, \cite{sethuraman2023nodags} reports a causal graph based on interventional genetic data in \cite{frangieh2021multimodal}, thus forming a complete benchmark for causal discovery.
However, the original papers of some datasets correspond to the dataset resource instead of the causal graph proposal paper.

Some datasets have different versions.
Instances include the Ecoli dataset versions, e.g., Ecoli70 and Ecoli100.
The Alzheimer has several versions, e.g., \cite{poldrack2015long}, \cite{shen2020challenges}, and several others.
We list different versions into one item.

Some datasets are updated regularly, and we list the latest version, e.g., the APM (Application performance monitoring program of AWS), and the General social survey.

Some papers claim that two datasets, namely, the Earthquake and the Survey, are practical.
However, we cannot find any realistic resource of these two datasets, and therefore, we neglect these two datasets in our list.

\subsection{Full list of the benchmarks and causal discovery papers}
\begin{enumerate}
\item Sachs\\
Original paper \cite{sachs2005causal}\\
Papers using it:              \cite{eulig2025toward,shen2025exploring,roy2025causal,shahverdikondori2024qwo,kang2026score,aglietti2023constrained,olko2023trust,annadani2023bayesdag,perry2022causal,dai2022independence,addanki2021collaborative,cundy2021bcd,wang2025federated,li2025strong,elahi2024adaptive,wang2024optimal,kaltenpoth2023nonlinear,rolland2022score,yi2025robustness,ke2022learning,zhang2023boosting,lippe2021efficient,wang2026robust,yin2024effective,liang2024multi,yang2024federated,guo2024fedcsl,bellot2024scores,yang2023reinforcement,ling2025hybrid,ling2025gradient,chen2024individual,niu2022learning,wang2021ordering}                                       
\item Child\\
Original paper           \cite{spiegelhalter1992learning}\\
Papers using it:              \cite{shen2025exploring,peyrard2020ladder,olko2023trust,wang2025federated,wang2024optimal,vashishtha2025causal,duong2025causal,ke2022learning,lippe2021efficient,guo2024fedcsl,zhang2023differentially,zhang2022residual,wang2022efficient,ling2025hybrid,guo2024sample,cui2022empirical}  
\item Alarm\\
Original paper           \cite{beinlich1989alarm}\\
Papers using it:              \cite{akbari2021recursive,roy2025causal,xie2024local,li2022hybrid,olko2023trust,wang2025federated,duong2025causal,lippe2021efficient,guo2024fedcsl,zhang2023differentially,zhang2022residual,wang2022efficient,zhang2021testing,ling2025hybrid,guo2024sample,cui2022empirical}                                          
\item Asia\\
Original paper            \cite{lauritzen1988local}\\
Papers using it:              \cite{shen2025exploring,roy2025causal,shahverdikondori2024qwo,olko2023trust,kocaoglu2023characterization,addanki2021collaborative,vashishtha2025causal,duong2025causal,ke2022learning,lippe2021efficient,bellot2024scores,zhang2023differentially,zhang2022residual}                                      
\item Insurance\\
Original paper       \cite{binder1997adaptive} \\
Papers using it:             \cite{mokhtarian2023novel,akbari2021recursive,feng2025iris,feng2025reliability,peyrard2020ladder,wang2025federated,guo2024fedcsl,zhang2022residual,wang2022efficient,zhang2021testing,guo2024sample,cui2022empirical}                                       
\item Cancer\\
Original paper          \cite{korb2010bayesian} \\
Papers using it:             \cite{feng2025iris,feng2025reliability,shahverdikondori2024qwo,peyrard2020ladder,olko2023trust,vashishtha2025causal,duong2025causal,lippe2021efficient,zhang2023differentially,zhang2022residual}                                         
\item Barley\\
Original paper          \cite{kristensen2002use}\\
Papers using it:             \cite{akbari2021recursive,wang2025federated,ling2025local,zhang2022residual,zhang2021testing,ling2025hybrid}                                          
\item Hailfinder\\
Original paper          \cite{abramson1996hailfinder}\\
Papers using it:              \cite{mokhtarian2023novel,akbari2021recursive,li2022hybrid,zhang2021testing,ling2025hybrid,cui2022empirical}                                           
\item Fmri hippocampus\\
Original paper            \cite{poldrack2015long}\\
Papers using it:              \cite{li2024federated,chen2024individual,zeng2021causal}                                         
\item Alzheimer\\
Original paper       \cite{petersen2010alzheimer,shen2020challenges}\\
Papers using it:              \cite{abdulaal2024causal,feng2025iris,vashishtha2025causal}                                         
\item Arctic sea ice\\
Original paper  \cite{huang2021benchmarking} \\
Papers using it:             \cite{abdulaal2024causal,feng2025reliability,kiciman2023causal}                                        
\item Diabetes\\
Original paper        \cite{long2022can} \\
Papers using it:              \cite{feng2025iris,feng2025reliability,lippe2021efficient}                                      
\item Ecoli70(100)\\
Original paper         \cite{schafer2005shrinkage} \\
Papers using it:             \cite{mokhtarian2023novel,akbari2021recursive,kang2026score,chen2021multi}                                          
\item Gene expression\\
Original paper         \cite{sethuraman2023nodags}\\
Papers using it:             \cite{guruswamy2026differentiable,xie2024local,li2025local,guo2024fedcsl,ling2025hybrid}                                           
\item Hepar2\\
Original paper         \cite{onisko2003probabilistic}\\
Papers using it:              \cite{mokhtarian2023novel,roy2025causal,li2022hybrid}                                         
\item Pigs\\
Original paper         \cite{fv1996introduction}\\
Papers using it:              \cite{lippe2021efficient,guo2024fedcsl,ling2025hybrid}                                          
\item Reged\\
Original paper         \cite{statnikov2015ultra}\\
Papers using it:              \cite{mian2021discovering,kaltenpoth2023nonlinear,guo2024sample}                                         
\item Andes\\
Original paper         \cite{conati1997line}\\
Papers using it:              \cite{xie2024local,li2022hybrid}                                         
\item Arth150\\
Original paper         \cite{opgen2007correlation}\\
Papers using it:             \cite{mokhtarian2023novel,akbari2021recursive,kang2026score}                                           
\item Auto mpg\\
Original paper         \cite{quinlan1993combining}\\
Papers using it:              \cite{eulig2025toward,shen2025exploring}                                          
\item Carpo\\
Original paper         \href{https://www.cs.huji.ac.il/w~galel/Repository/Datasets/carpo/carpo.htm}{Link}\\
Papers using it:             \cite{mokhtarian2023novel,akbari2021recursive}                                          
\item HK stock\\
Original paper         \cite{huang2020causal} \\
Papers using it:             \cite{li2024federated,cai2023causal}                                           
\item Link\\
Original paper         \cite{jensen1999blocking} \\
Papers using it:             \cite{wang2022efficient,ling2025hybrid}                                           
\item Mildew\\
Original paper         \cite{jensen1996midas} \\
Papers using it:            \cite{xie2024local,li2022hybrid}                                           
\item Neuropain\\
Original paper         \cite{tu2019neuropathic} \\
Papers using it:             \cite{liu2024discovery,feng2025iris,kiciman2023causal,vashishtha2025causal}                                           
\item Obesity\\
Original paper         \cite{long2022can} \\
Papers using it:             \cite{feng2025iris,feng2025reliability}                                           
\item Syntren\\
Original paper         \cite{van2006syntren} \\
Papers using it:             \cite{dhir2024continuous,ling2025hybrid}                                          
\item WIN95PTS\\
Original paper         \cite{heckerman1995learning}\\
Papers using it:              \cite{xie2024local,wang2022efficient,zhang2021testing}                                           
\item CE-Tueb\\
Original paper         \cite{mooij2016distinguishing}\\
Papers using it:              \cite{dhir2023bivariate}                                           
\item CE-cha\\
Original paper         \cite{guyon2019evaluation}\\
Papers using it:              \cite{dhir2023bivariate}                                          
\item Cognition and Aging in the Chronic Fatigue Syndrome\\
Original paper         \cite{heins2013process}\\
Papers using it:              \cite{qiao2024identification}                                           
\item DWDClimate\\
Original paper         \cite{mooij2016distinguishing}\\
Papers using it:              \cite{shen2025exploring}                                          
\item MAGIC-IRRI\\
Original paper         \cite{scutari2016bayesian}\\
Papers using it:              \cite{kang2026score}                                           
\item MAGIC-NIAB\\
Original paper         \cite{scutari2014multiple}\\
Papers using it:              \cite{kang2026score}                                           
\item New York Times\\
Original paper         \href{https://www.kaggle.com/datasets/BidecInnovations/stock-price-and-news-realted-to-it}{Link}\\
Papers using it:              \cite{liu2024discovery}                                           
\item Pittsburgh Bridges\\
Original paper         \cite{reich1989incremental}\\
Papers using it:              \cite{ni2022bivariate}                                           
\item Categorical Cause-Effect Pairs\\
Original paper         \cite{ni2022bivariate}\\
Papers using it:              \cite{ni2022bivariate}                                           
\item Abalone\\
Original paper         \cite{warwick1994population}\\
Papers using it:              \cite{ni2022bivariate}                                           
\item Alcohol\\
Original paper         \cite{long2022can}\\
Papers using it:              \cite{feng2025reliability}                                           
\item Algibra I\\
Original paper         \href{https://pslcdatashop.web.cmu.edu/KDDCup/}{Link}\\
Papers using it:              \cite{yu2024causal}                                           
\item APM\\
Original paper         \href{https://docs.aws.amazon.com/AmazonCloudWatch/latest/monitoring/CloudWatch-Application-Monitoring-Intro.html}{Link}\\
Papers using it:              \cite{eulig2025toward}                                           
\item Apple gastronome\\
Original paper         \cite{liu2024discovery}\\
Papers using it:              \cite{feng2025iris}                                           
\item Big five\\
Original paper         \href{https://openpsychometrics.org/}{Link}\\
Papers using it:              \cite{dai2025latent,dong2024versatile}                                          
\item Brain tumor\\
Original paper         \href{https://github.com/SartajBhuvaji/Brain-Tumor-Classification-Using-Deep-Learning-Algorithms}{Link}\\
Papers using it:              \cite{liu2024discovery}
\item Chemical\\
Original paper         \cite{ke2021systematic}\\
Papers using it:              \cite{zhao2025curious}                                          
\item (no?)Chemistry image\\
Original paper         \cite{schafer2005shrinkage}\\
Papers using it:              tba                                         
\item climatic analysis\\
Original paper         \cite{compo2011twentieth}\\
Papers using it:              \cite{liu2024discovery}                                           
\item COVID-19\\
Original paper         \href{https://bayesian-ai.eecs.qmul.ac.uk/bayesys/}{Link}\\
Papers using it:              \cite{wang2025llm,vashishtha2025causal}                                          
\item Credit\\
Original paper         \cite{credit_approval_27}\\
Papers using it:              \cite{li2022hybrid}                                           
\item dream\\
Original paper         \cite{kalainathan2019causal}\\
Papers using it:              \cite{roy2025causal}                                          
\item football\\
Original paper        \href{https://www.kaggle.com/datasets/secareanualin/football-events}{Link}\\
Papers using it:              \cite{qiao2024causal}                                           
\item G7\\
Original paper         \cite{demirer2018estimating}\\
Papers using it:              \cite{jalaldoust2022causal}                                           
\item General social survey\\
Original paper         \href{https://gss.norc.org/get-the-data.html}{Link}\\
Papers using it:              \cite{li2025local}                                         
\item IHDP\\
Original paper         \cite{hill2011bayesian}\\
Papers using it:              \cite{ashman2023causal}                                           
\item Lucas\\
Original paper         \cite{lucas2004bayesian}\\
Papers using it:              \cite{roy2025causal}                                           
\item Magnetic\\
Original paper         \cite{hwang2024fine}\\
Papers using it:              \cite{zhao2025curious}                                           
\item (tba)Micro24\\
Original paper         \cite{schafer2005shrinkage}\\
Papers using it:              tba                                          
\item (tba)Micro25\\
Original paper         \cite{schafer2005shrinkage}\\
Papers using it:              tba                                          
\item Munin\\
Original paper         \cite{andreassen1989munin} \\
Papers using it:             \cite{dong2025dcilp,wang2022efficient}                                                              
\item Pacific walker circulation\\
Original paper         \cite{runge2019detecting} \\
Papers using it:             \cite{liu2023causal}                                           
\item Pathfinder\\
Original paper         \cite{heckerman1992toward}\\
Papers using it:              \cite{zhang2021testing}                                           
\item Pharmacokinetics\\
Original paper         \cite{grzegorzewski2021pk}\\
Papers using it:              \cite{li2024causal}                                          
\item Physics\\
Original paper         \cite{lee2024incorporating} \\
Papers using it:             \cite{kang2026score}                                           
\item Protein\\
Original paper         \href{https://www.cs.huji.ac.il/w~galel/Repository/}{Link}\\
Papers using it:              \cite{peyrard2020ladder}                                           
\item Respiratory disease\\
Original paper         \cite{long2022can} \\
Papers using it:             \cite{feng2025iris}                                           
\item Sangiovese\\
Original paper         \cite{magrini2017conditional}\\
Papers using it:              \cite{abdulaal2024causal}                                           
\item SCM\\
Original paper         \cite{ferro2015use}\\
Papers using it:              \cite{aglietti2023constrained,sussex2022model}                                           
\item Sepsis\\
Original paper         \cite{matot2001definition}\\
Papers using it:              \cite{liu2024causal}                                          
\item T\"ubingen\\
Original paper         \cite{mooij2016distinguishing}\\
Papers using it:              \cite{ni2022bivariate,meier2026additive,tran2025identifying,kiciman2023causal,tu2022optimal,yin2024effective,ton2021meta}                                          
\item Water\\
Original paper         \cite{jensen1989expert}\\
Papers using it:             \cite{mokhtarian2023novel}                                           
\item World value survey\\
Original paper         \href{https://www.worldvaluessurvey.org/WVSDocumentationWV7.jsp}{Link}\\
Papers using it:              \cite{dai2025latent}                                          
\item Yahoo\\
Original paper: \cite{janzing2010telling}\\
Papers using it: \cite{zeng2021causal}                                          
\end{enumerate}

\subsection{LLM-based causal discovery papers}\label{sec:LLM_papers}

\begin{figure}[t]
    \centering
    \includegraphics[width=\textwidth]{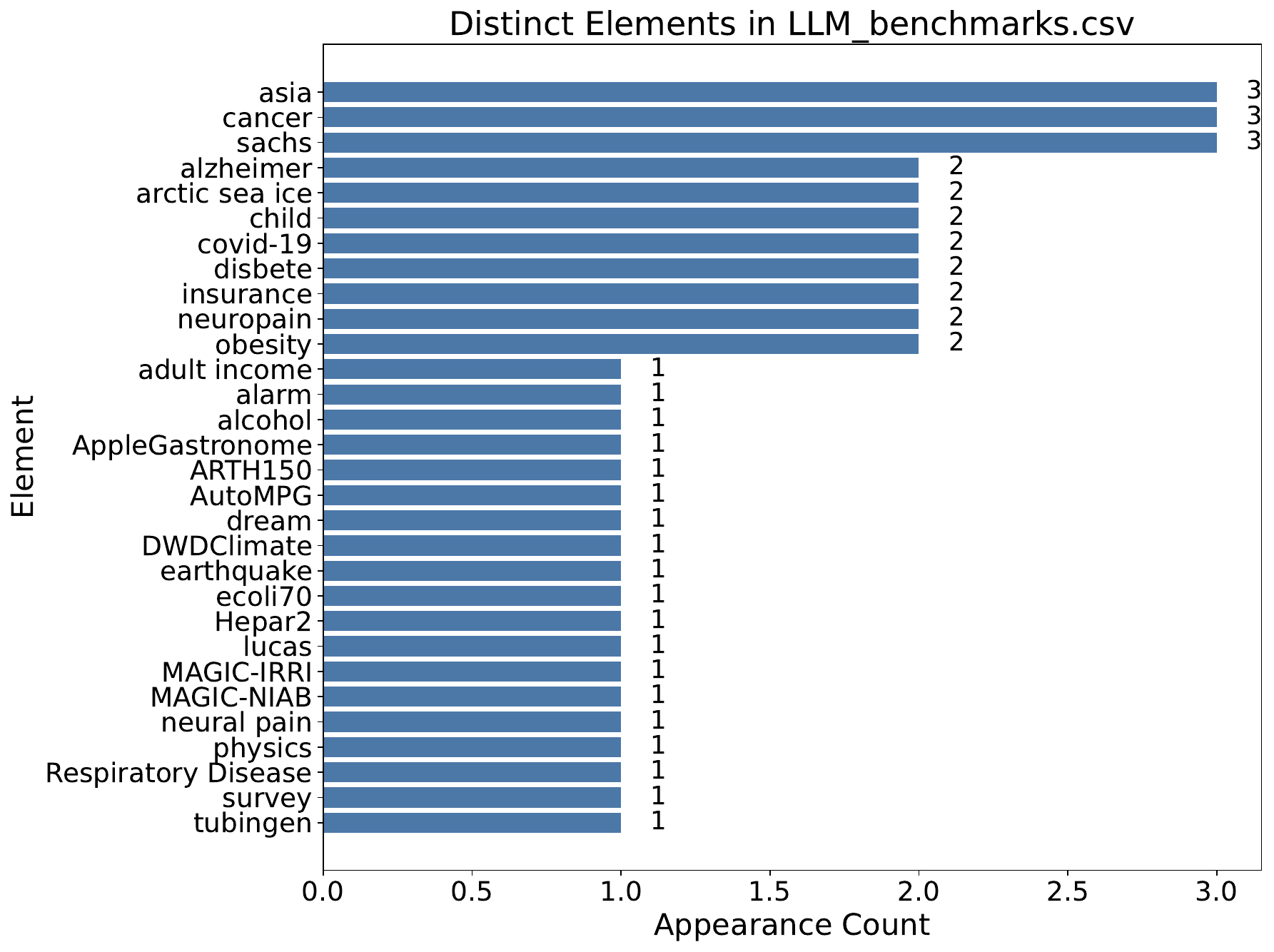}
    \caption{Benchmark numbers used in LLM-based papers.}
    \label{fig:llm_benchmarks}
\end{figure}

Figure \ref{fig:llm_benchmarks} shows the benchmarks used in our surveyed LLM-based causal discovery works, with the number of papers that use each benchmark.

\subsection{LLM-based causal discovery papers benchmarks on arXiv}\label{sec:arxiv_benchmark}

\begin{figure}[t]
    \centering
    \includegraphics[width=\textwidth]{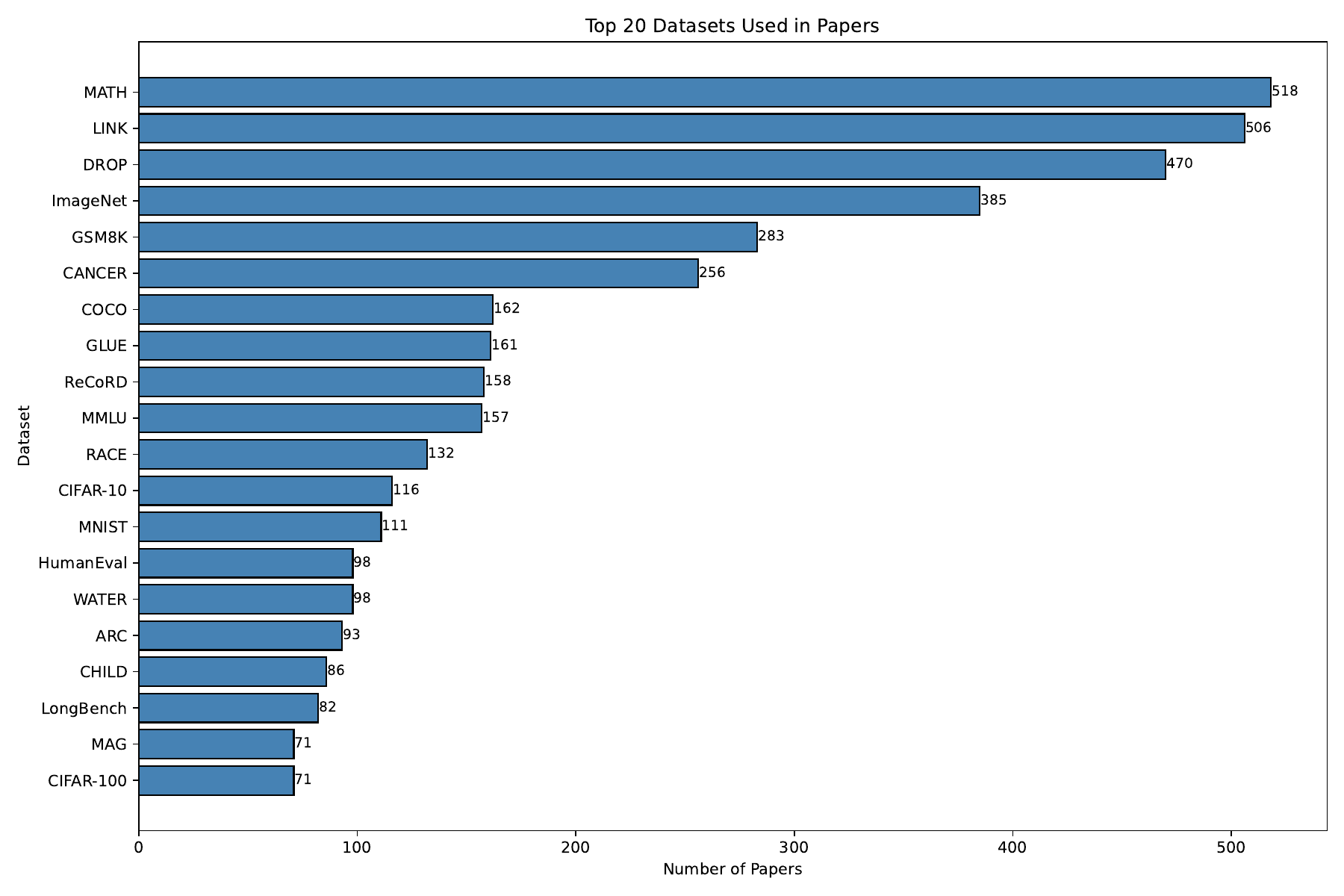}
    \caption{Top 20 benchmarks used in arXiv LLM-based causal discovery papers.}
    \label{fig:arxiv_benchmark}
\end{figure}

Figure \ref{fig:arxiv_benchmark} shows the top 20 benchmarks that are mostly used in recent LLM-based causal discovery works on arXiv.
The numbers are the papers that use each benchmark.

\section{Additional information of the benchmark evaluation process}\label{sec:app:process}
\subsection{Conditional association relationship extraction algorithm}
Algorithm \ref{algo:association} is used to extract the conditional association queries from a causal graph for a given variable pair.

\begin{algorithm}[h]
\caption{Conditional association relationship extraction}
\begin{algorithmic}[1]
\REQUIRE{Causal graph $\g = \langle \V, \E \rangle$, and target variable pair $X,Y\in \V$}
\IF{$X$ and $Y$ are adjacent in $\g$}
\STATE{Return: $X$ and $Y$ are always correlated.}
\ELSE
\STATE{$\mathbf{P}\leftarrow$ all collider-free paths between $X$ and $Y$.}
\IF{$\mathbf{P}=\emptyset$}
\STATE{Return: $X$ and $Y$ are mutually independent.}
\ELSE
\STATE{$SepSet\leftarrow \emptyset$.}
\STATE{$\mathbf{D}\leftarrow \{\V'\subseteq \V\setminus\{X,Y\}\mid \forall P\in \mathbf{P}, P\text{ contains at least one variable in }\V'\}$.}
\FOR{$\Z\in \mathbf{D}$}
\IF{$\nexists \Z'\subset \Z$ such that $\Z'\in \mathbf{D}$}
\STATE{$SepSet.append(\Z)$}
\ENDIF
\ENDFOR
\STATE{Return: [$X$ and $Y$ are mutually independent conditioned on $\Z$ for $\Z$ in $SepSet$]}
\ENDIF
\ENDIF
\end{algorithmic}
\label{algo:association}
\end{algorithm}

\subsection{Analysed paper distribution}\label{sec:app:paper_count}

As shown in Table \ref{tab:benchmark_relationship_counts}, a large part of the retrieved papers do not contain sufficient information to answer the conditional association query, i.e., return answer ``Unknown''.
This high rate of ``Unknown'' papers is because of the paper search mechanism of the scientific databases: the keyword matching rule is inefficient to retrieve target papers.
Note that the Unknown papers may be higher than the actual number of the irrelevant papers.
The reason is that a given variable pair may have multiple conditional association queries, since the pair can be d-separated by multiple variable sets.
A paper may contain information to verify the consistency of one query, but for the other queries, the answer is ``Unknown''.

\begin{figure}[h]
    \centering
    \includegraphics[width=\textwidth]{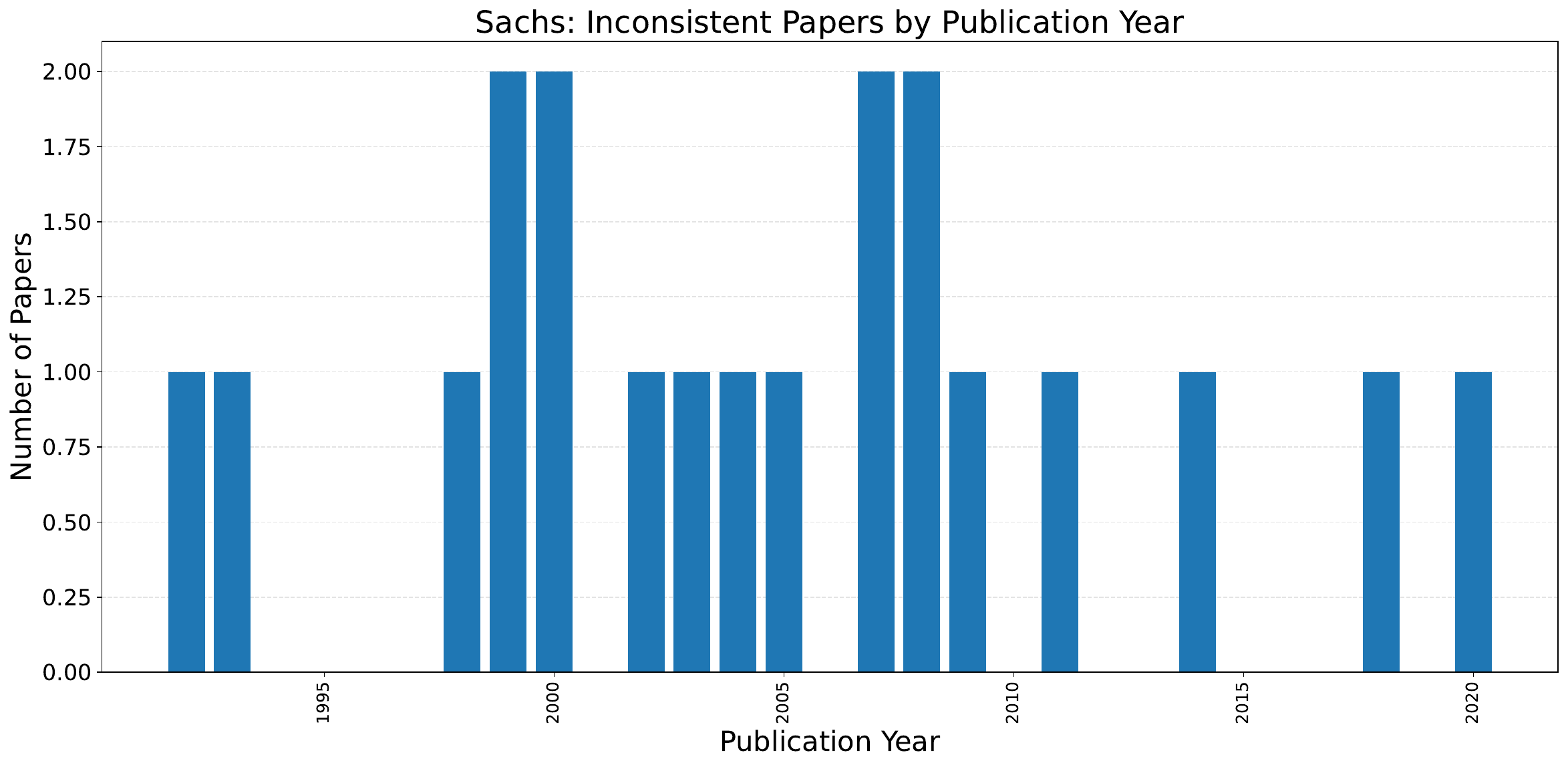}
    \caption{Inconsistent paper number per year of Sachs.}
    \label{fig:sachs_inconsistent}
\end{figure}

\begin{figure}[h]
    \centering
    \includegraphics[width=\textwidth]{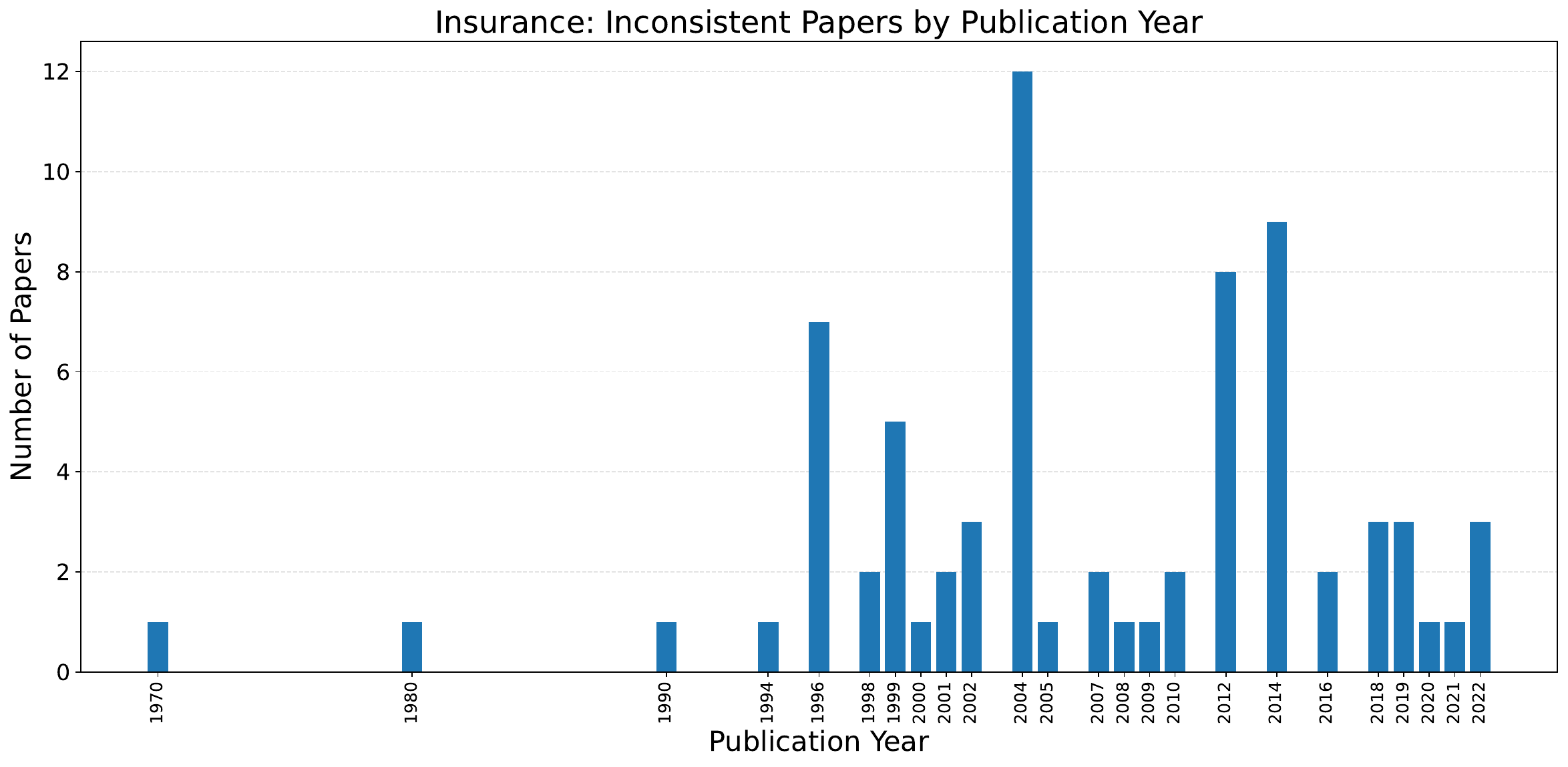}
    \caption{Inconsistent paper number per year of Insurance.}
    \label{fig:insurance_inconsistent}
\end{figure}

\begin{figure}[h]
    \centering
    \includegraphics[width=\textwidth]{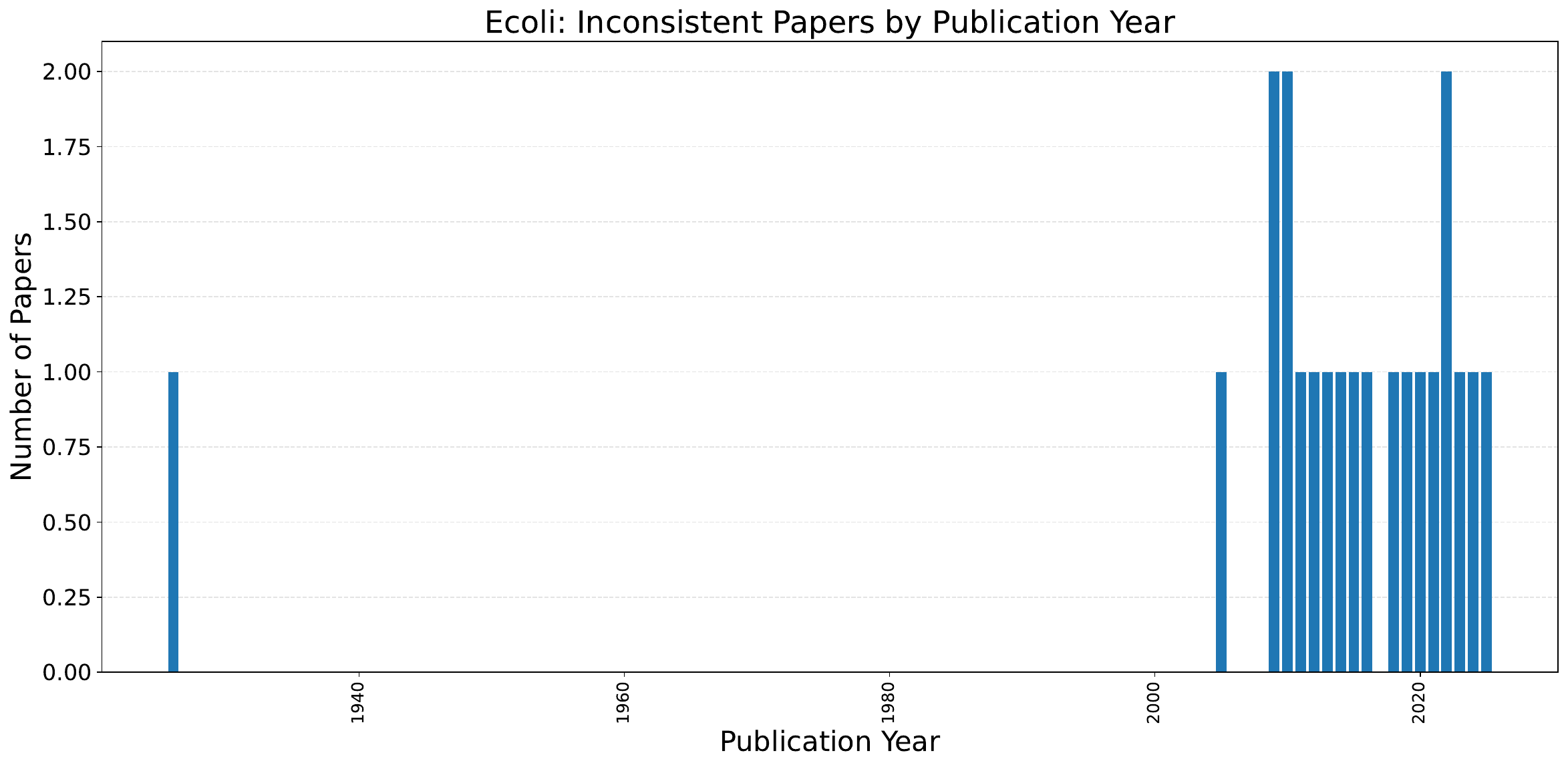}
    \caption{Inconsistent paper number per year of Ecoli.}
    \label{fig:ecoli_inconsistent}
\end{figure}

\begin{figure}[h]
    \centering
    \includegraphics[width=\textwidth]{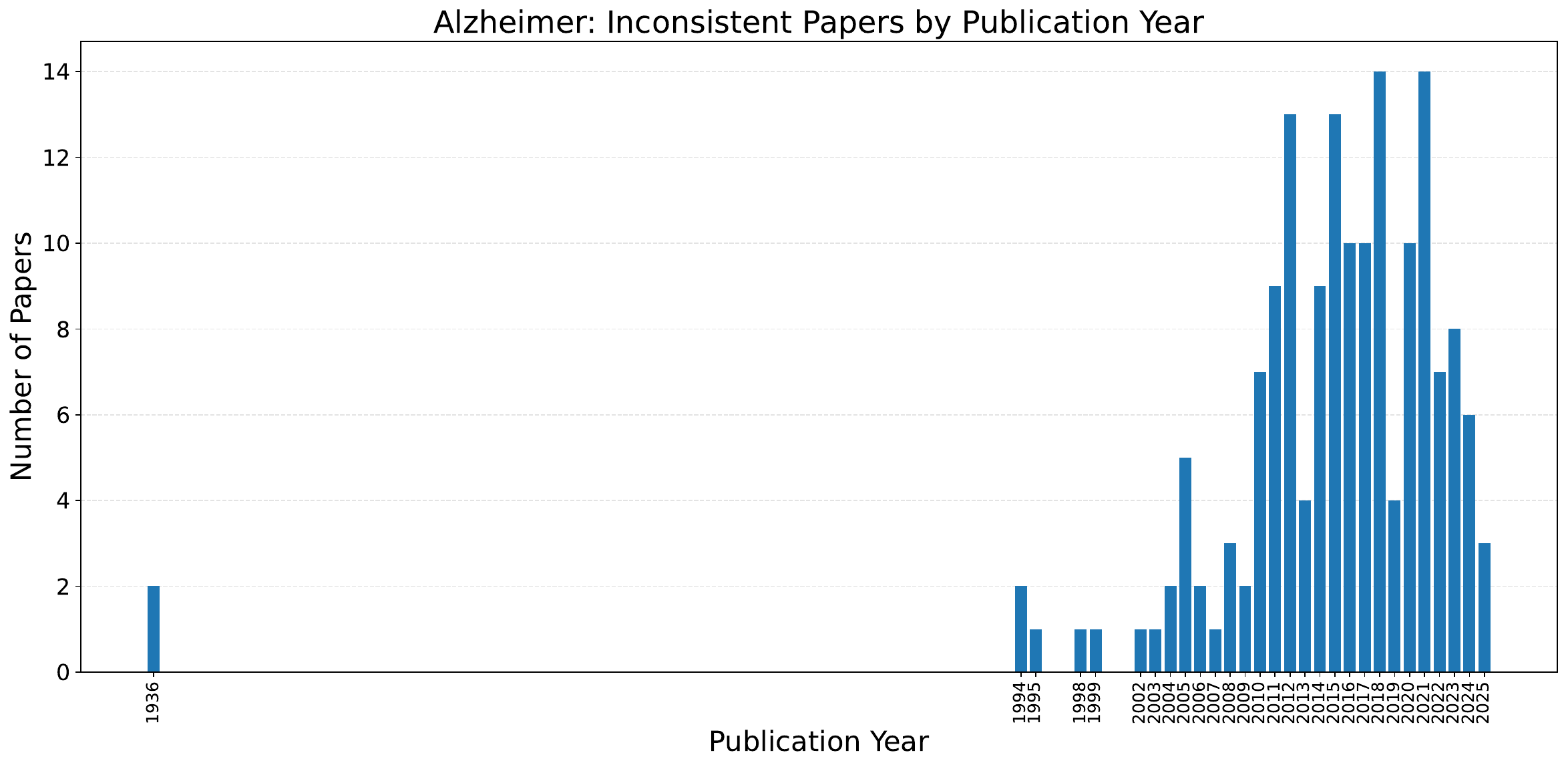}
    \caption{Inconsistent paper number per year of Alzheimer.}
    \label{fig:alzheimer_inconsistent}
\end{figure}

\begin{figure}[h]
    \centering
    \includegraphics[width=\textwidth]{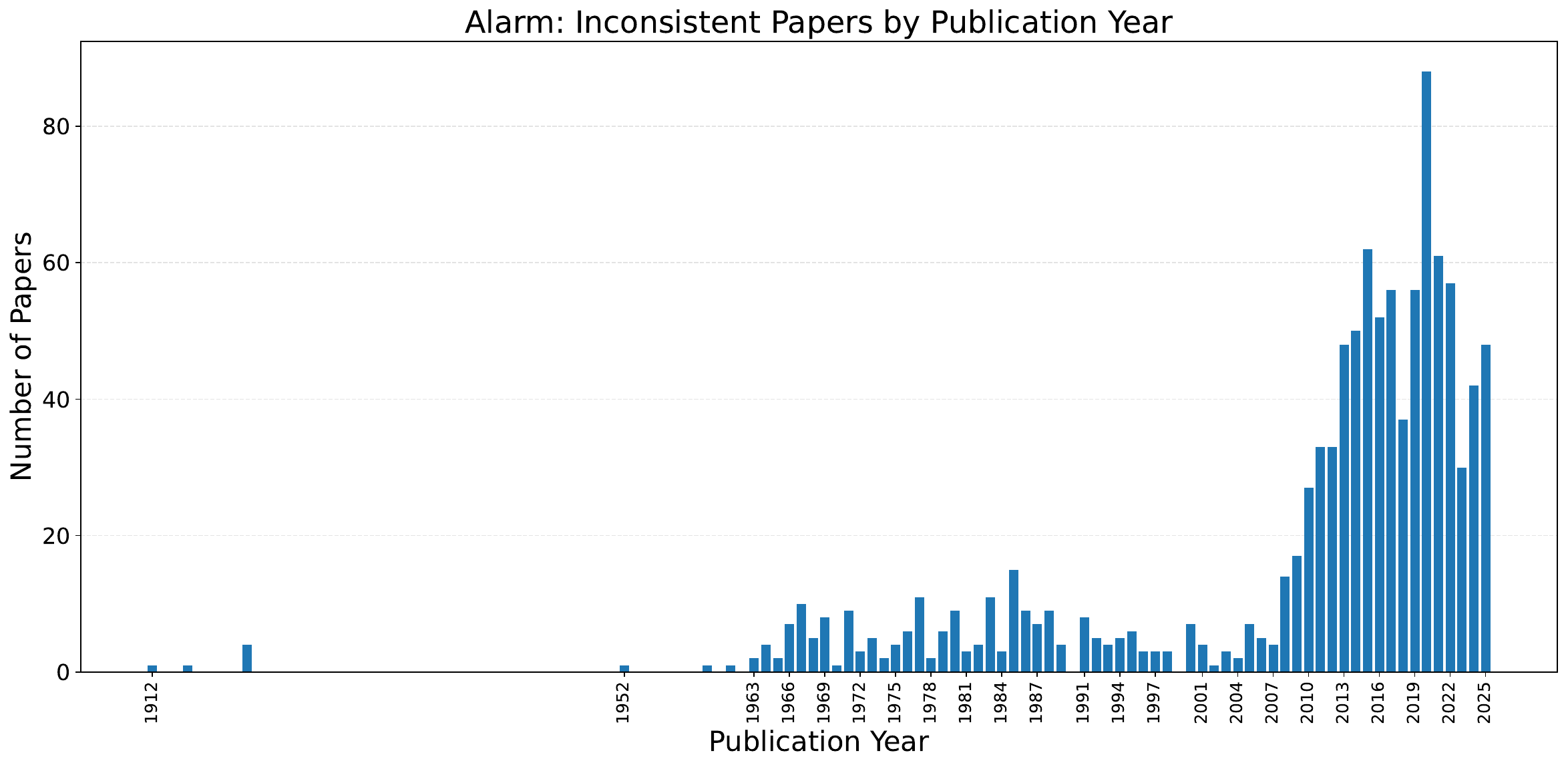}
    \caption{Inconsistent paper number per year of Alarm.}
    \label{fig:alarm_inconsistent}
\end{figure}

\begin{figure}[h]
    \centering
    \includegraphics[width=\textwidth]{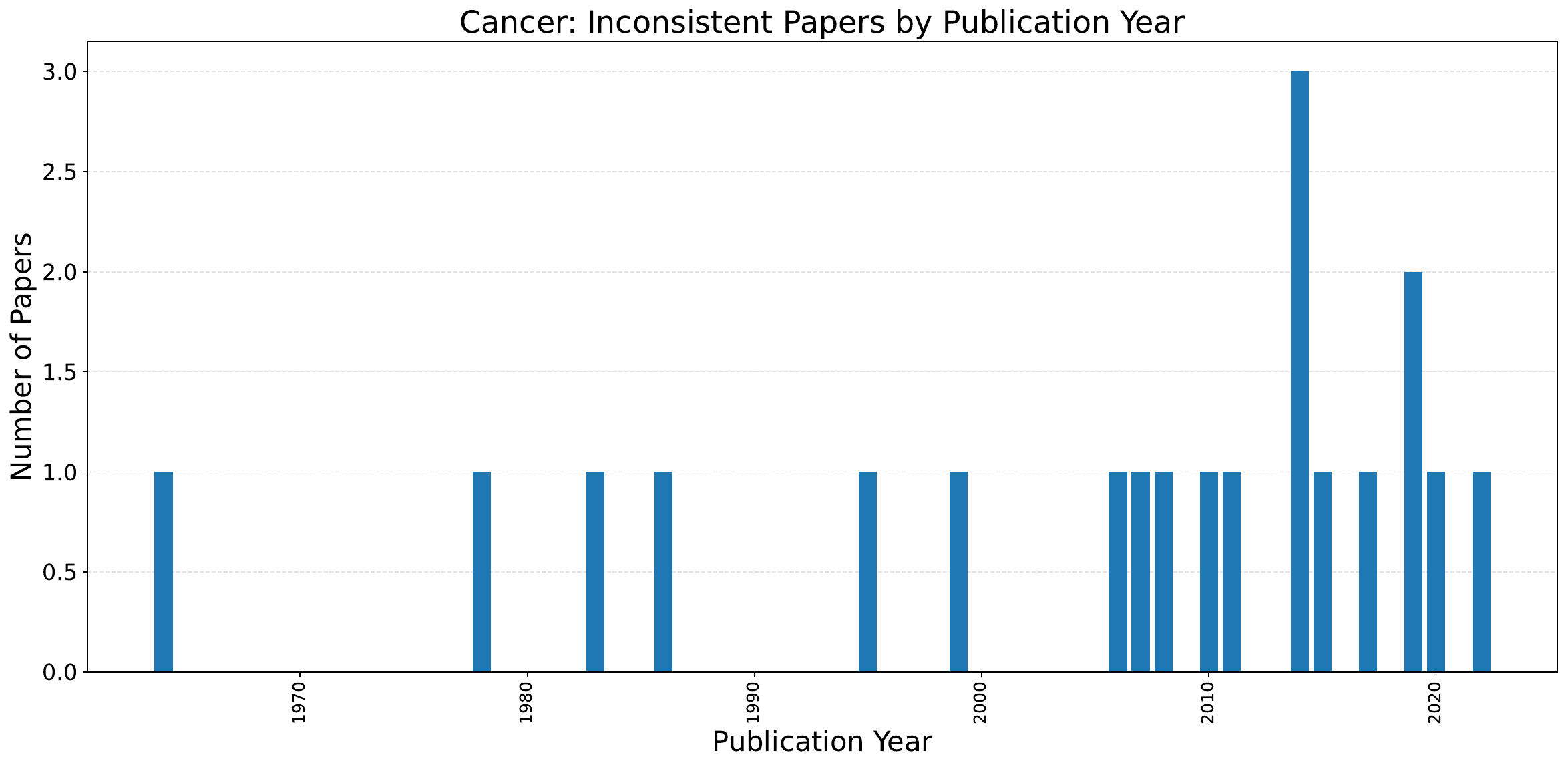}
    \caption{Inconsistent paper number per year of Cancer.}
    \label{fig:cancer_inconsistent}
\end{figure}

\begin{figure}[h]
    \centering
    \includegraphics[width=\textwidth]{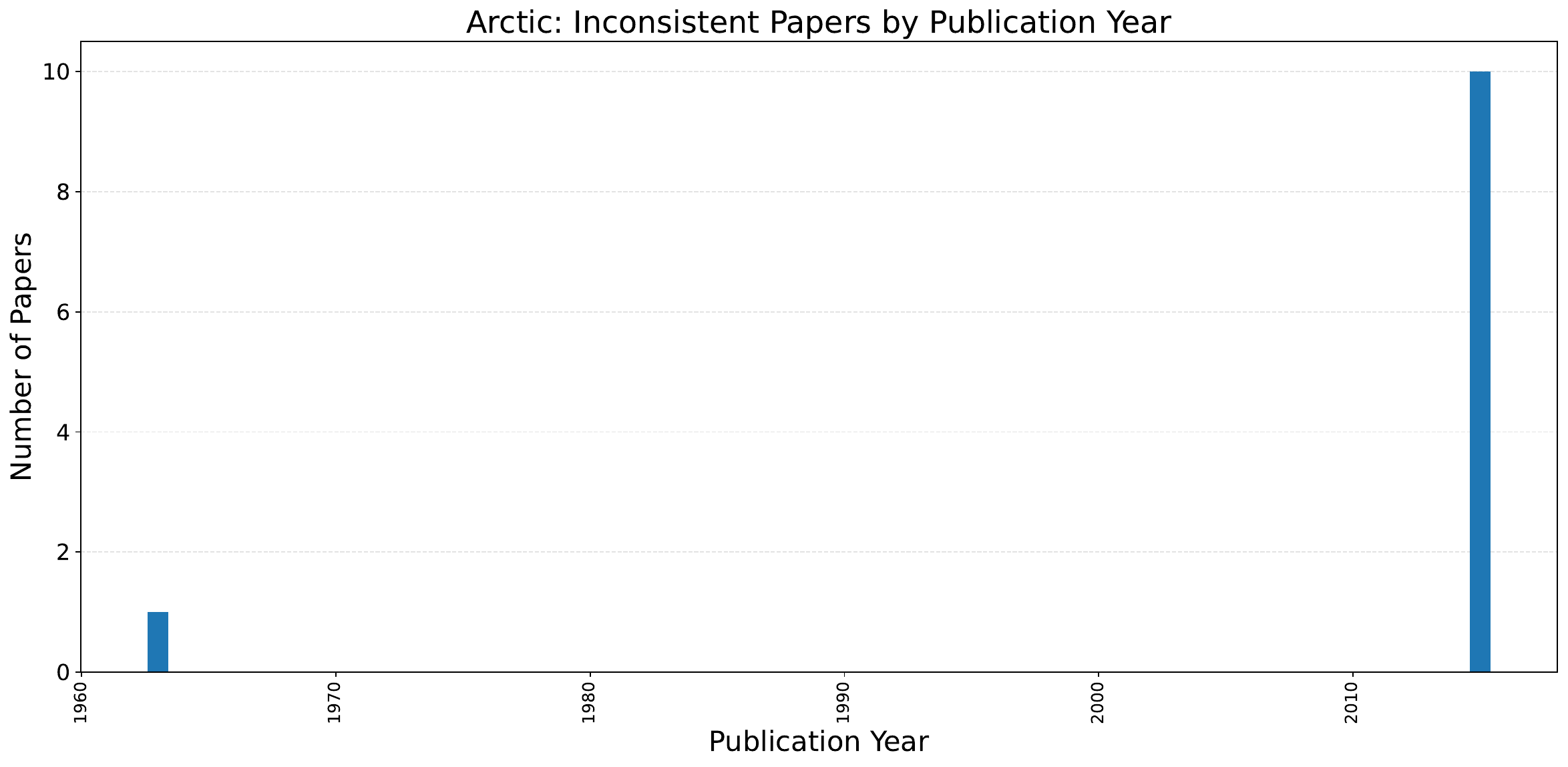}
    \caption{Inconsistent paper number per year of Arctic Sea Ice.}
    \label{fig/arctic_inconsistent}
\end{figure}

\begin{figure}[h]
    \centering
    \includegraphics[width=\textwidth]{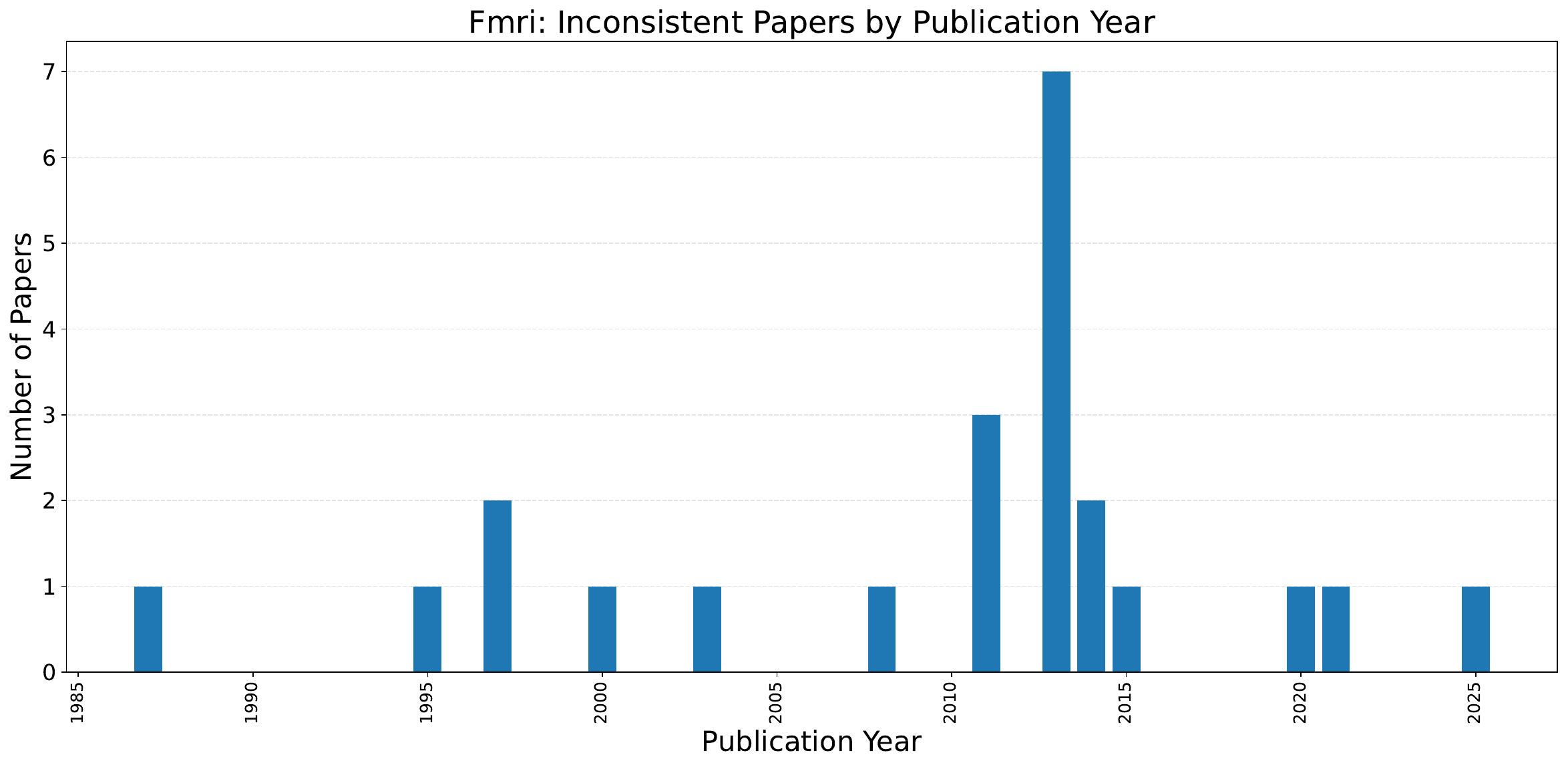}
    \caption{Inconsistent paper number per year of Fmri.}
    \label{fig:fmri_inconsistent}
\end{figure}

\begin{figure}[h]
    \centering
    \includegraphics[width=\textwidth]{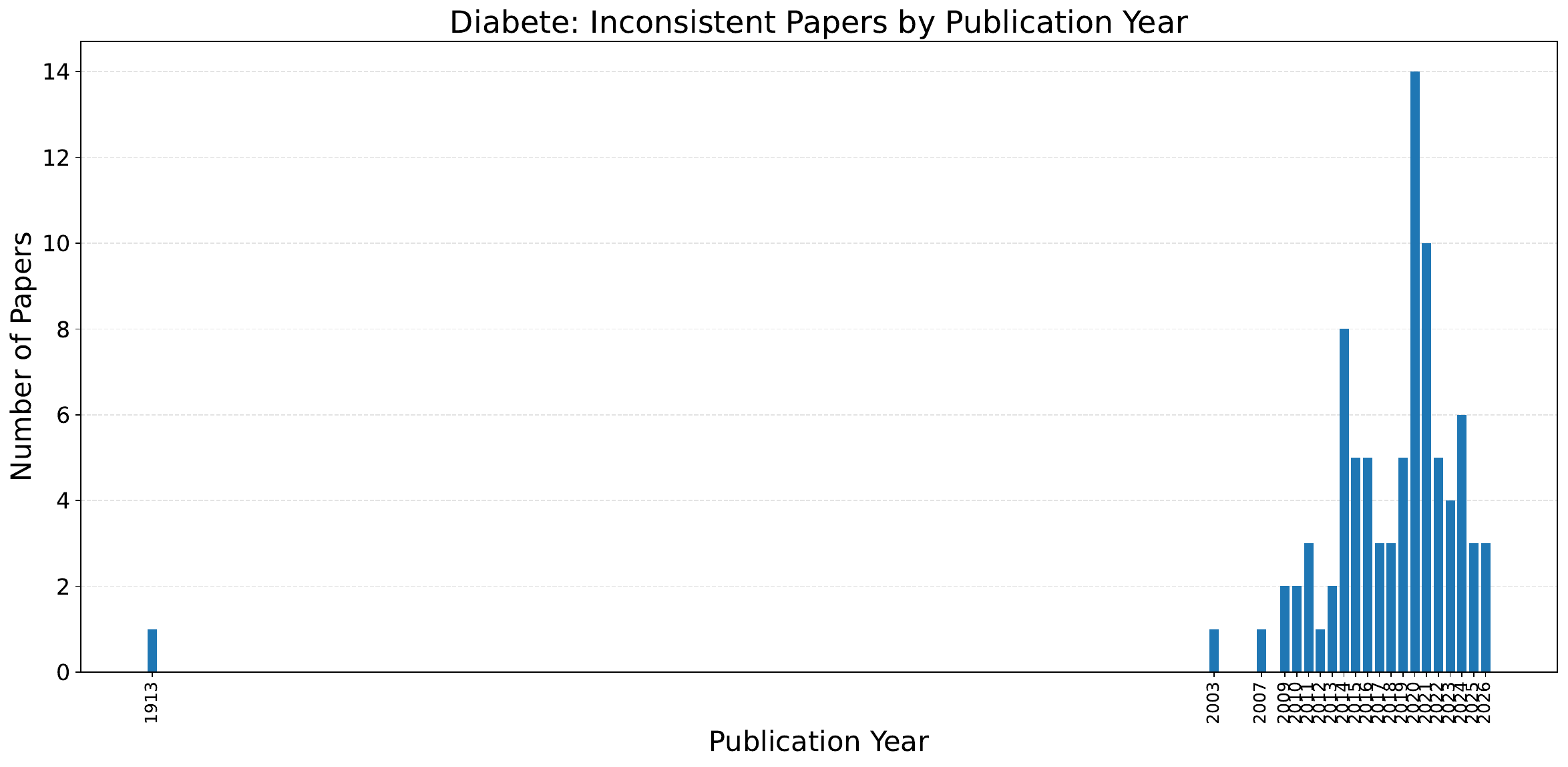}
    \caption{Inconsistent paper number per year of Diabetes.}
    \label{fig:diabetes_inconsistent}
\end{figure}

\begin{figure}[h]
    \centering
    \includegraphics[width=\textwidth]{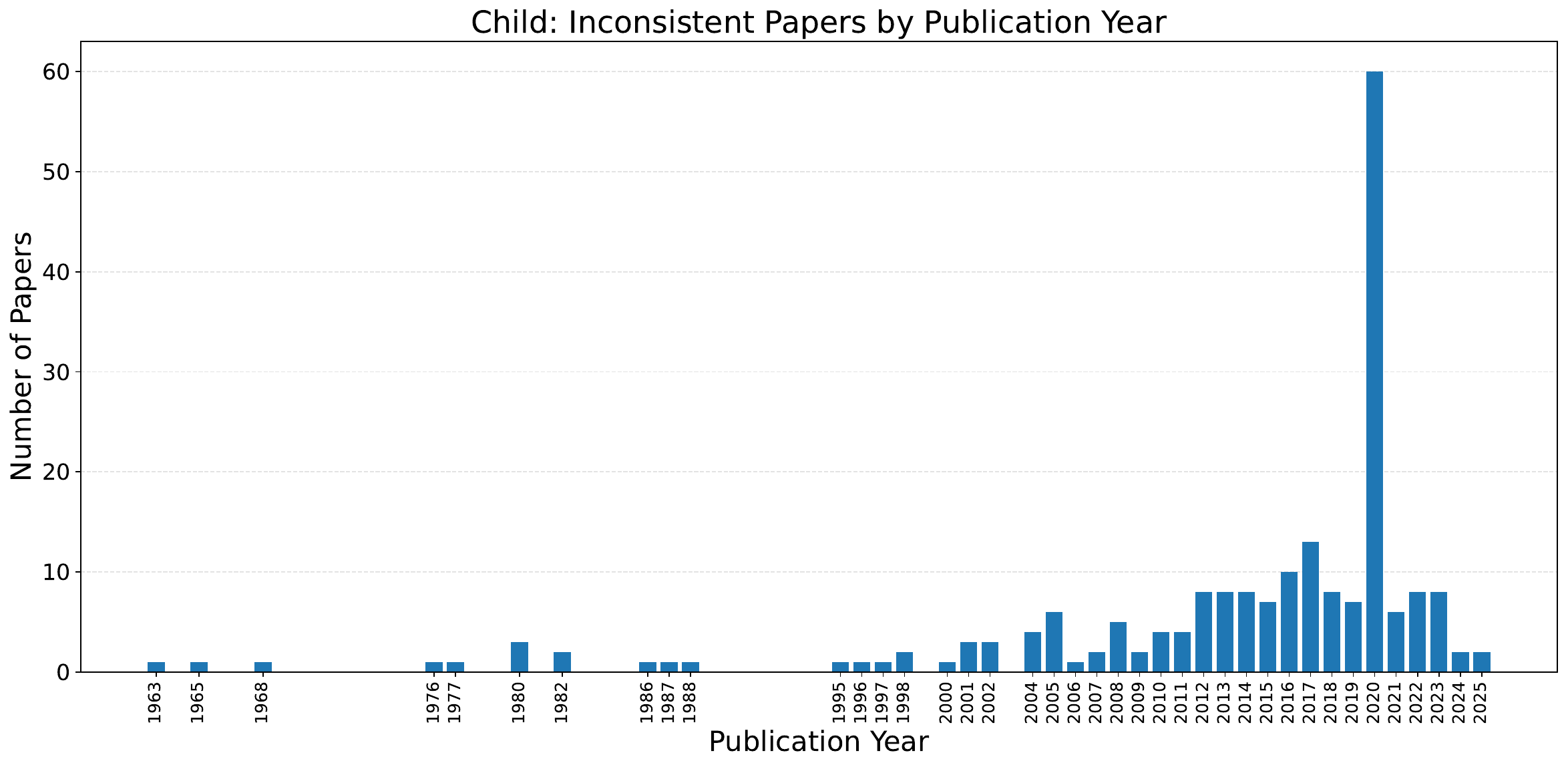}
    \caption{Inconsistent paper number per year of Child.}
    \label{fig:child_inconsistent}
\end{figure}

\begin{figure}[h]
    \centering
    \includegraphics[width=\textwidth]{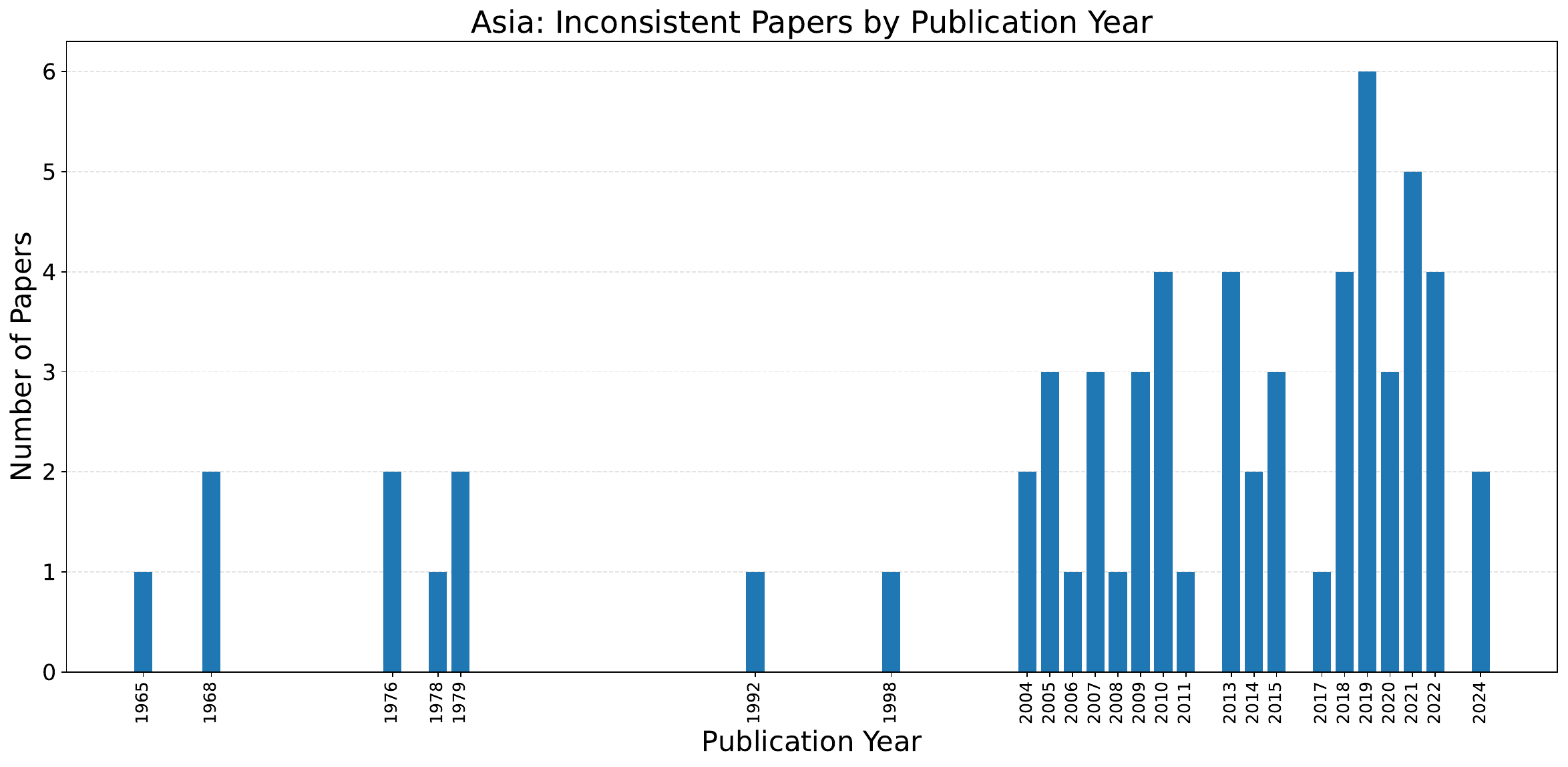}
    \caption{Inconsistent paper number per year of Asia.}
    \label{fig:asia_inconsistent}
\end{figure}

Figures \ref{fig:sachs_inconsistent} to \ref{fig:asia_inconsistent} show the inconsistent paper number per year for all evaluated benchmarks.

\begin{table}[htbp]
\centering
\begin{tabular}{|l|r|r|r|r|}
\hline
Benchmark & Correlated & Independent & Inconsistent & Unknown \\
\hline
alarm & 1262 & 904 & 1054 & 23667 \\
\hline
alzheimer & 150 & 286 & 167 & 5105 \\
\hline
arctic sea ice & 14 & 12 & 11 & 422 \\
\hline
asia & 74 & 33 & 62 & 1333 \\
\hline
cancer & 34 & 16 & 20 & 648 \\
\hline
child & 212 & 173 & 208 & 4490 \\
\hline
diabetes & 189 & 3 & 87 & 252 \\
\hline
ecoli & 19 & 51 & 23 & 6981 \\
\hline
fmri & 19 & 35 & 23 & 2103 \\
\hline
insurance & 70 & 177 & 78 & 7709 \\
\hline
sachs & 14 & 61 & 20 & 3797 \\
\hline
\end{tabular}
\caption{Paper number distribution for each evaluated benchmarks.}
\label{tab:benchmark_relationship_counts}
\end{table}


\subsection{LLM-based paper verifier accuracy}\label{sec:app:accuracy}
A pipeline component that potentially brings high noise is the LLM-based consistency verifier which employs the LLMs to analyse papers and decide the papers' consistency with the conditional associations.
We randomly sample 100 retrieved papers, and manually verify the correctness of the LLM-based consistency verifier's answers.
The accuracy achieves 90\%, i.e., the LLM-based consistency verifier correctly analyses 90 out of the 100 randomly sampled papers.
See evaluation details, including sampled papers, LLM responses, and manual labels, on \href{https://iamyoezy.github.io/LLM_CA_verifier_evaluation/}{this website}.


\subsection{LLM prompt}
You are a scientific research assistant analyzing a paper on \{pair\_info\}.
Your task is to determine if this paper contains data analysis that can answer the \{query\}.
You must return ONLY a JSON object with the structure specified in the user's instructions.

First, you need to thoroughly read the following "Paper content", and then to determine whether the paper contains data analysis information based on the following principles.
You are an experienced researcher and good at analysing research papers. Think step by step to answer the \{query\}.

"has\_data\_analysis determination principles":\\
A paper contains data analysis information if:\\
1. It has a concrete description of the data used in the study or data collection methods.\\
2. It has a detailed illustration on the statistical methods or machine learning based methods, used to analyse the data.\\
3. It contains detailed reports, for example, the analysis results and significance (for example, p values), on the above analysis methods and data.\\
4. *IMPORTANT* Do check whether the paper introduces any numerical dataset and implements corresponding analysis on the dataset. A review/survey paper does not contain data analysis.\\

If the paper contains data analysis information, then you need to determine whether the data analysis can answer the \{query\}.
Typically, we are interested in the statistical correlation between the main factors of \{pair\_info\}, possibly conditioned on a set of other factors (called covariates).
Then, you can answer the query \{query\} by the following principles:

1. The query may ask whether two factors are correlated or independent conditioned on some other factors. This query is to ask the statistical correlational relationship between the factor pair conditioned on other factors.\\
2. The statistical correlation between the factor pair can be revealed by statistical tests or machine learning results based on data described in the paper.
3. We can conclude that the pair of factors are correlated if the paper shows a significant statistical test result for a correlation test (for example Pearson test or regression), otherwise we can conclude that the factor pair are mutually independent.
4. When the query is asking the statistical correlation between the factor pair conditioned on a set of factors, we consider the statistical correlation of the conditional density.
5. Sometimes, the conditioned factors are called covariates. A research usually consider adjusting statistical analysis by covariates, and under such processes, we can confirm the work investigates the correlation (correlated or independent) of the factor pair conditioned on the covariates.
6. The correlation of the conditional density means: if we fix the conditioned factors constant, are the pair of factors mutually independent or correlated?
7. In the research paper, the conditional correlation may be considered as: in the data collection experiments (e.g., random trial experiments), researchers may control the value of some factors (the covariates), and collect corresponding data for the main factor pair. In such cases, we can obtain reliable conclusion about the conditional correlation.
8. Reason indication of covariates. For example, factor Smoking may be strongly correlated with age. If a data sample has an age under 18, it may indicate that the sample does not smoke.
9. When it is possible to reason indication of factor, do not implement the reasoning for \{pair\_info\}.
10. Two factors are mutually independent marginally if the marginal density of the two factors are statistically independent, but the independence may not hold when conditioned on other factors.
11. If the query asks whether two factors are correlated directly, you do not need to consider the conditional factors, that is, the covariates.

IMPORTANTLY, you need to get clear about the following points:
1. Check and remember the covariates mentioned in the paper's method description.
2. Be clear about the main factors.
3. The covariates may be correlated with or independent from the main factors. This should not influence our investigation on the correlation between the main factors.
4. Be careful whether the factors studied in the paper are really those in \{pair\_info\} and query.

"Required information"
1. Answer the query with a single choice among: "Correlated", "Independent", and "Unknown".
2. A brief thinking process of how you obtain the conclusion.
3. Reference sentences from the original paper for the conclusion.

Paper content (possibly truncated):
```
\{paper\_text\}
```

[IMPORTANT and MUST] Return ONLY a JSON object with this structure:
\{\{
  "results":\{\{
    "has\_data\_analysis": true or false,
    "answer of the query": Independent or Correlated or Unknown,
    "brief thinking process": the brief reasoning process of how you obtain the answer,
    "reference": the reference contents from the original paper corresponding to your answer.
    \}\}
\}\}
"""

\subsection{Additional results}

\begin{figure}[t]
    \centering
    \includegraphics[width=\textwidth]{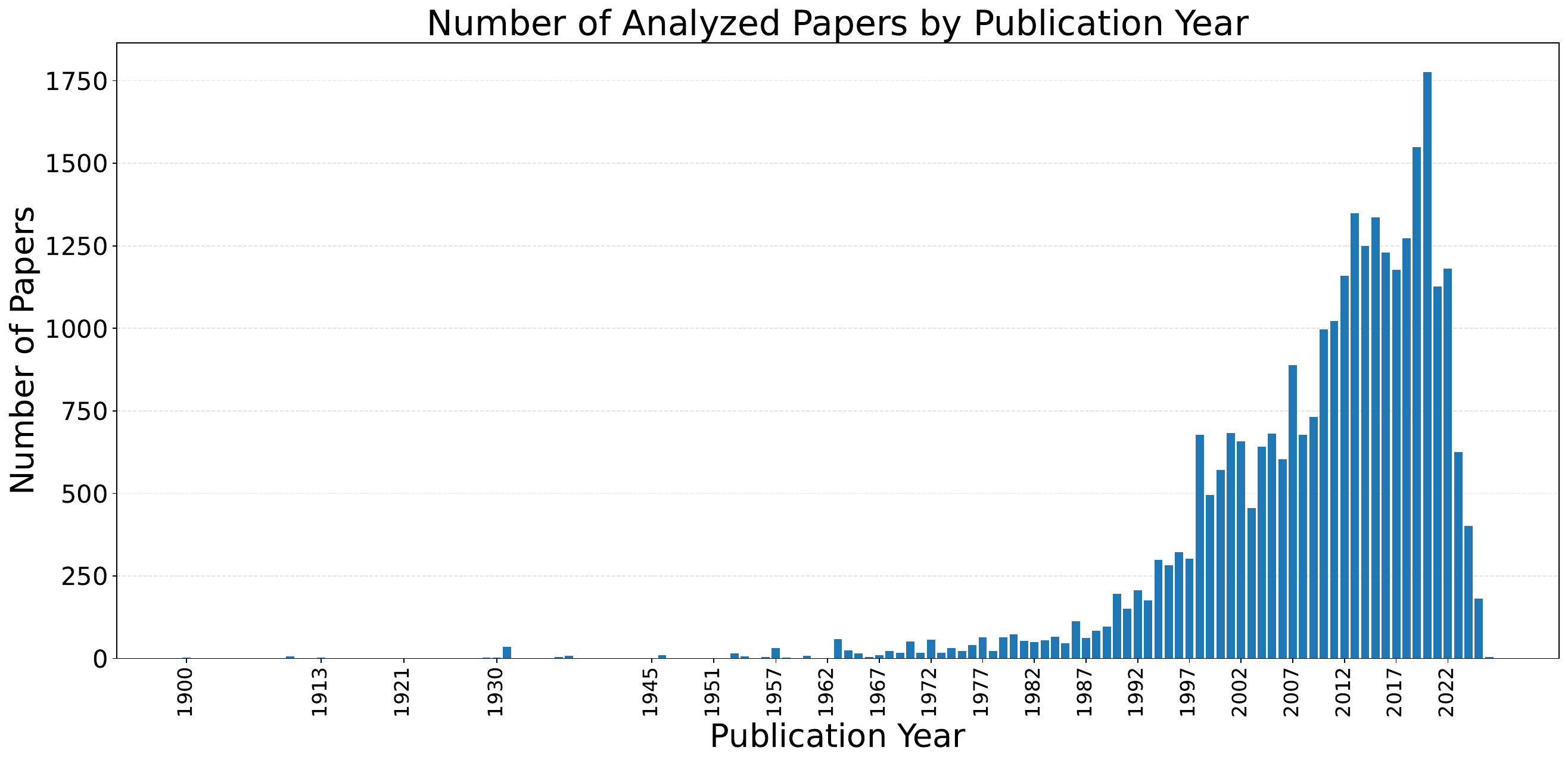}
    \caption{Retrieved papers by year.}
    \label{fig:paper_by_year}
\end{figure}

\begin{figure}[t]
    \centering
    \includegraphics[width=\textwidth]{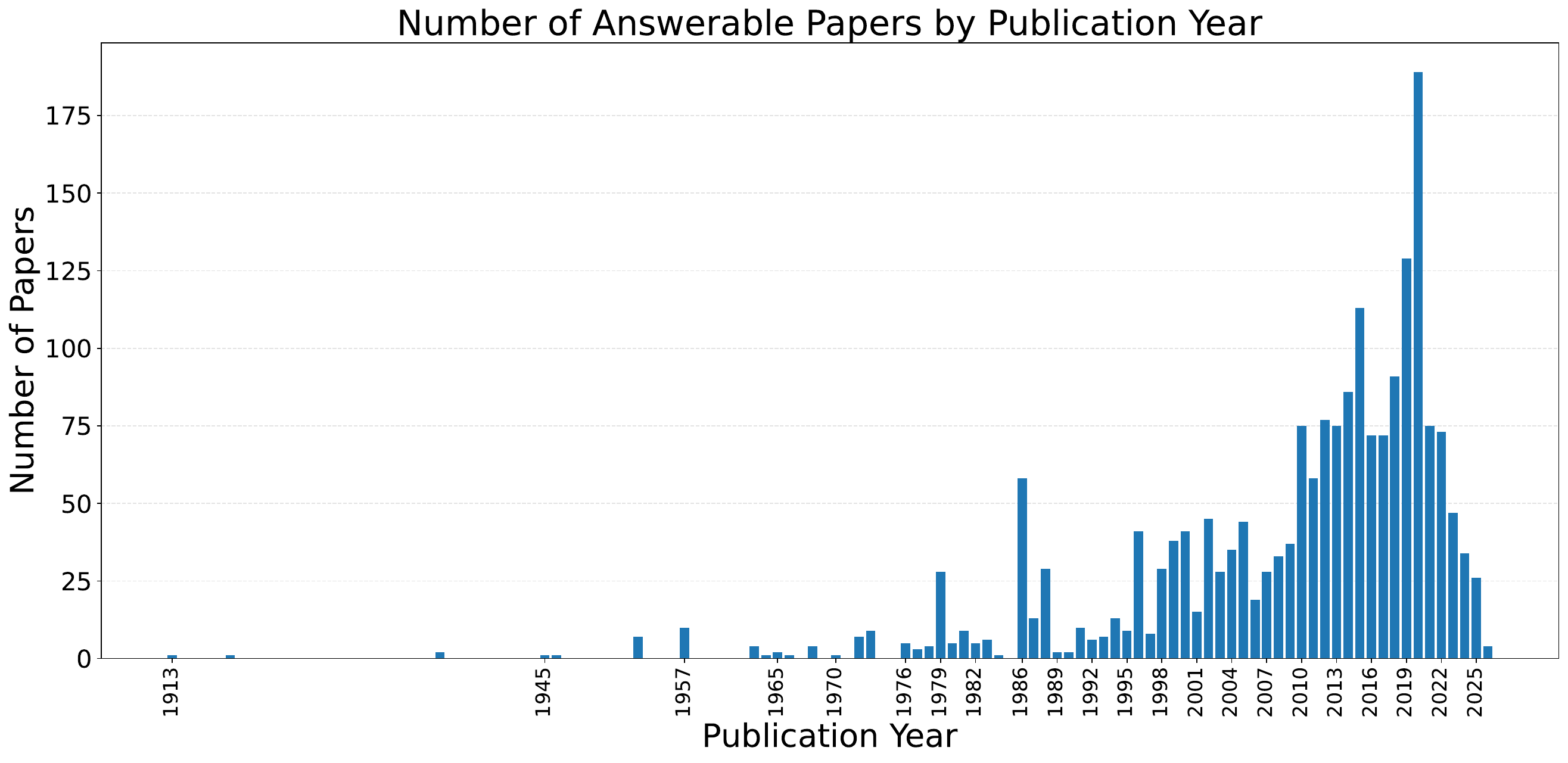}
    \caption{Retrieved papers that contain relevant information by year.}
    \label{fig:relevant_paper_by_year}
\end{figure}

Figures \ref{fig:paper_by_year} and \ref{fig:relevant_paper_by_year} show the numbers of retrieved papers and relevant papers per year, respectively, for all evaluated benchmarks.

\end{document}